\RequirePackage{fix-cm}
\documentclass[smallextended]{svjour3}       %
\smartqed  %
\usepackage{graphicx}
\usepackage{microtype}

\usepackage{url}
\usepackage{latexsym}

\usepackage{tabularx}
\usepackage{tcolorbox}
\usepackage{color, colortbl}
\usepackage{graphicx}
\usepackage{tabularx}
\usepackage{soul}
\usepackage{pbox}
\usepackage{multirow}
\usepackage{float}
\usepackage{multirow}
\usepackage{url}
\usepackage[show]{inotes}
\usepackage{booktabs}
\usepackage{color}
\usepackage{breakurl}
\usepackage{listings}
\usepackage{multicol}
\usepackage{graphics}
\usepackage{hyperref}
\usepackage[ruled,vlined]{algorithm2e}
\usepackage{xstring}
\usepackage[inline]{enumitem}
\usepackage{epstopdf}
\usepackage[utf8]{inputenc}
\usepackage{amssymb}%
\usepackage{pifont}%
\newcommand{\xmark}{\ding{53}}%
\usepackage[normalem]{ulem}
\useunder{\uline}{\ul}{}
\usepackage{hyperref}
\usepackage{xstring}
\usepackage{caption}
\usepackage{subcaption}

\usepackage{color}
\usepackage{enumitem}
\usepackage{comment}

\usepackage{soul,xcolor}

\def\threedigits#1{%
  \number#1}
  
\def\makeheadbox{\relax}

\begin{document}

\title{
Robust Training of Social Media Image Classification Models for Rapid Disaster Response

}

\titlerunning{Robust Training of Social Media Image Classification Models}        %

\author{Firoj Alam \and
        Tanvirul Alam \and
        Muhammad Imran \and
        Ferda Ofli
}

\institute{F. Alam, M. Imran, F. Ofli \at
              Qatar Computing Research Institute, HBKU, Doha, Qatar \\
              \email{\{fialam, mimran, fofli\}@hbku.edu.qa}           %
           \and
           T. Alam \at
              BJIT Limited, Dhaka, Bangladesh\\
              \email{tanvirul.alam@bjitgroup.com}
}

\date{Received: date / Accepted: date}

\maketitle

\begin{abstract}
Images shared on social media help crisis managers gain situational awareness and assess incurred damages, among other response tasks. As the volume and velocity of such content are typically high, real-time image classification has become an urgent need for a faster disaster response. Recent advances in computer vision and deep neural networks have enabled the development of models for real-time image classification for a number of tasks, including detecting crisis incidents, filtering irrelevant images, classifying images into specific humanitarian categories, and assessing the severity of the damage. To develop robust real-time models, it is necessary to understand the capability of the publicly available pre-trained models for these tasks, which remains to be under-explored in the crisis informatics literature. In this study, we address such limitations by investigating ten different network architectures for four different tasks using the largest publicly available datasets for these tasks. We also explore various data augmentation strategies, semi-supervised techniques, and a multitask learning setup. In our extensive experiments, we achieve promising results.

\keywords{Social media image classification \and Crisis informatics \and Humanitarian tasks \and Disaster response \and Real-time classification}
\end{abstract}

\section{Introduction}
\label{sec:introduction}

Social media is widely used during natural or human-induced disasters to disseminate information and obtain valuable insights quickly. People post content (i.e., through different modalities such as text, image, and video) on social media to ask for help, to offer support, to identify urgent needs, or to share their feelings. Such information is helpful for humanitarian organizations to plan and launch relief operations. As the volume and velocity of the content are significantly high, it is crucial to have real-time systems to process social media content to facilitate rapid response automatically. 
There has been a surge of research works in this domain in the past couple of years. The focus has been to analyze social media data and develop computational models using varying modalities to extract actionable information. Among different modalities (e.g., text and image), more focus has been given to textual content analysis compared to imagery content (see \cite{imran2015processing,said2019natural,Imran:IPM20} for comprehensive surveys). However, many past research works have demonstrated that images shared on social media during a disaster event can also assist humanitarian organizations. For example, Nguyen et al.~\cite{nguyen17damage} use images shared on Twitter to assess the severity of the infrastructure damage, and Mouzannar et al.~\cite{Mouzannar2018} focus on identifying damages in infrastructure as well as environmental elements.

\begin{figure}[!htb]
        \centering
        \begin{subfigure}{.9\textwidth}
            \includegraphics[width=\textwidth]{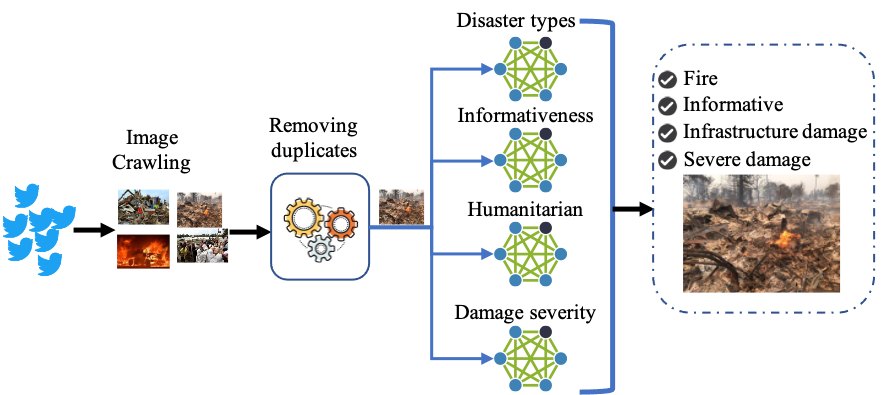}
            \caption{Disaster image classification pipeline.}
            \label{fig:use_case_pipeline}
        \end{subfigure}\\
        \begin{subfigure}{.9\linewidth}
                \includegraphics[width=\textwidth]{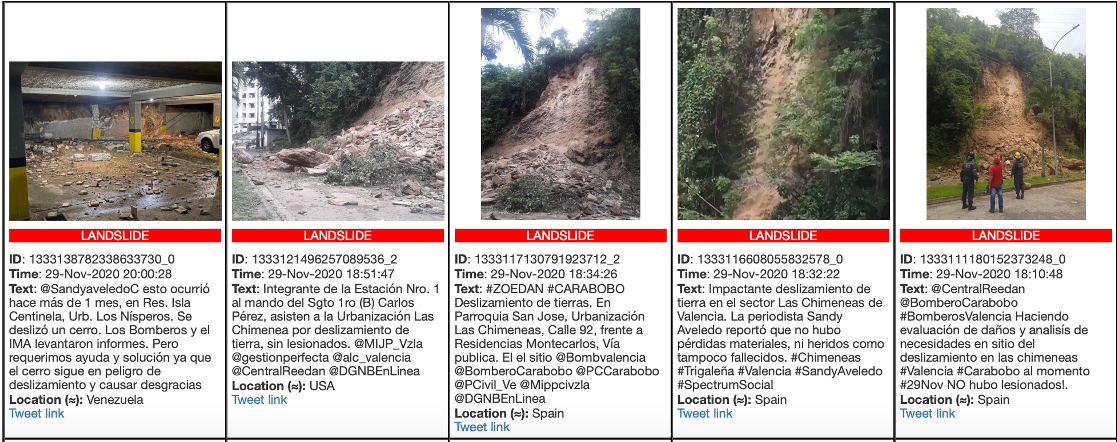}
                \caption{Event detection use case showing landslide images.}
                \label{fig:use_case_landslide_event_detection}
        \end{subfigure}
    \caption{Disaster image classification pipeline that demonstrate a real use case -- landslide image classification.}
    \label{fig:pipeline_and_use_case}
\end{figure}

For a clear understanding we provide an example pipeline in Figure~\ref{fig:use_case_pipeline} which demonstrates how different disaster-related image classification models can be used in real-time for information categorization. As presented in the figure, the four different classification tasks such as {\em (i)} disaster types, {\em (ii)} informativeness, {\em (iii)} humanitarian, and {\em (iv)} damage severity assessment, can significantly help crisis responders during disaster events. For example, disaster type classification model can be used for real-time event detection as shown in Figure~\ref{fig:use_case_landslide_event_detection}. Similarly, the informativeness model can be used to filter non-informative images, the humanitarian model can be used to discover fine-grained categories, and the damage severity model can be used to assess the impact of the disaster. Current literature reports either one or two tasks using one or two network architectures. Another limitation is that there have been limited datasets for disaster-related image classification. Very recently the study by Alam et al.~\cite{FAlam:ASONAM20} developed a \textit{benchmark dataset},\footnote{We refer to this dataset as \textit{Crisis Benchmark Dataset} throughout the paper.} which is consolidated from existing publicly available resources. The development process of this dataset consists of data curation from different existing sources, development of new data for new tasks, creating non-overlapping\footnote{Duplicate images are identified between test and training sets and moved from the test set to the training set.} training, development, and test sets. The reported benchmark dataset targeted the four tasks as shown in Figure~\ref{fig:use_case_pipeline}.

In this study, we build upon \cite{FAlam:ASONAM20} and address the aforementioned limitations by posing the following Research Questions (RQs):
\begin{itemize}
    \itemsep0em
    \item \textbf{RQ1:} Can data consolidation help? 
    \item \textbf{RQ2:} Among various neural network architectures with pre-trained weights, which one is more suitable for different downstream disaster-related image classification tasks? 
    \item \textbf{RQ3:} Does data augmentation or semi-supervised learning help to improve the performance?
    \item \textbf{RQ4:} Is multitask learning an ideal solution to reduce computational complexity when there is need to make predictions for multiple tasks simultaneously?
\end{itemize}

To understand the benefits of data consolidation (\textit{RQ1}), we extended the work by Alam et al.~\cite{FAlam:ASONAM20} with more in-depth analysis. Our motivation for \textit{RQ2} is that there has been significant progress in neural network architectures for image processing in the past few years; however, they have not been widely explored in the \textit{crisis informatics}\footnote{\url{https://en.wikipedia.org/wiki/Disaster_informatics}} domain for disaster response tasks. Hence, we investigated %
several neural network architectures for different disaster-related image classification tasks. 
Since augmentation and self-training-based techniques~\cite{cubuk2020randaugment,lee2013pseudo} have shown success to yield a more generalized model and sometimes improve the performance, we posed \textit{RQ3} and investigated them for the mentioned tasks. 
For the real-time social media image classification tasks shown in Figure~\ref{fig:pipeline_and_use_case}, it is necessary to run the mentioned models in sequence or parallel for the same input image. 
Running multiple models can be prohibitively expensive when there is a need to analyze many social media images in real-time. Having a single model for dealing with multiple tasks can significantly alleviate the computational complexity. Hence, we posed \textit{RQ4} to instigate research in this direction.
The \textit{Crisis Benchmark Dataset} has not been originally developed for multitask learning setup. However, the related metadata information (e.g., image ids) are available, and we utilized such information to create data splits for multitask learning while trying to maintain the same training, development, and test splits. As our experiment shows, this is challenging due to the incomplete labels for different tasks (see more details in Section \ref{ssec:exp_multitask_learning}).

To summarize, our contributions in this study are as follows:
\begin{itemize}
    \itemsep0em
    \item We present more detailed results highlighting the benefit of data consolidation. 
    \item We address four tasks using several state-of-the-art neural network architectures on different data splits. 
    \item We investigate various data augmentation techniques and show that model generalization improves with data augmentation.  
    \item We explore semi-supervised learning and multitask learning to have a single model while addressing multiple tasks. Based on the findings, we provide research directions for future studies. 
    \item We also provide insights using Gradient-weighted Class Activation Mapping~\cite{selvaraju2017grad} to demonstrate what class-specific discriminative properties are learned by the networks.
\end{itemize}

The rest of the paper is organized as follows. Section~\ref{sec:related_works} provides a brief overview of the existing work. Section~\ref{sec:tasks} introduces the tasks and describes the datasets used in this study. Section~\ref{sec:experiments} explains the experiments, Section~\ref{sec:results} presents the results, and Section~\ref{sec:discussions} provides a discussion. Finally, we conclude the paper in Section~\ref{sec:conclutions}.

\section{Related Work}
\label{sec:related_works}

\subsection{Social Media Images for Disaster Response}
The studies on image processing in the crisis informatics domain are relatively few compared to the studies on analyzing textual content for humanitarian aid.\footnote{\url{https://en.wikipedia.org/wiki/Humanitarian_aid}} With recent successes of deep learning for image classification, research works have started to use social media images for humanitarian aid. The importance of imagery content on social media for disaster response tasks has been reported in many studies~\cite{petersinvestigating,daly2016mining,chen2013understanding,nguyen2017automatic,nguyen17damage,alam17demo,alam2019SocialMedia}. 
For instance, the analysis of flood images has been studied in \cite{petersinvestigating}, in which the authors reported that the existence of images with the relevant textual content is more informative. Similarly, the study by Daly and Thom~\cite{daly2016mining} analyzed fire event images, which are extracted from social media data. Their findings suggest that images with geotagged information are helpful to locate the fire-affected areas. 

The analysis of imagery content shared on social media has recently been explored using deep learning techniques for damage assessment purposes. Most of these studies categorize the severity of damage into discrete levels~\cite{nguyen2017automatic,nguyen17damage,alam17demo} whereas others quantify the damage severity as a continuous-valued index~\cite{nia2017building,li2018localizing}. 
Other related work include data scarcity issue by employing more sophisticated models such as adversarial networks~\cite{li2019identifying,pouyanfar2019unconstrained}, disaster image retrieval \cite{ahmad2017convolutional}, image classification in the context of bush fire emergency \cite{10.3389/frobt.2016.00054}, flooding photo screening system \cite{ning2020prototyping}, sentiment analysis from disaster image \cite{hassan2019sentiment}, monitoring natural disasters using satellite images \cite{ahmad2017jord}, and flood detection using visual features \cite{Jony2019FloodDI}.

\subsection{Real-time Systems}
Recently, Alam et al.~\cite{alam2019SocialMedia} presented an image processing pipeline to extract meaningful information from social media images during a crisis situation, which has been developed using deep learning-based techniques. Their image processing pipeline includes collecting images, removing duplicates, filtering irrelevant images, and finally classifying them with damage severity. 
Such a system has been used during several disaster events, and one such example is the deployment during Hurricane Dorian, reported in \cite{imran2020rapid}. The system has been deployed for 13 days, and it collected around $\sim$280K images. These images are then automatically classified and used by a volunteer response organization, Montgomery County Maryland Community Emergency Response Team (MCCERT).
Another example use case is the early detection of disaster-related damage to cultural heritage~\cite{10.1145/3383314}.

\subsection{Multimodality (Image and Text)}
The exploration of multimodality has also received attention in the research community \cite{agarwalcrisis,abavisani2020multimodal}. In \cite{agarwalcrisis}, authors explore different fusion strategies for multimodal learning. Similarly, in \cite{abavisani2020multimodal} a cross-attention-based network is exploited for multimodal fusion. The study in \cite{huang2019visual} reports a multimodal system for flood image detection, which achieves a precision of 87.4\% in a balance test set. In another study, the authors propose a similar multimodal system for on-topic vs. off-topic social media post classification and report an accuracy of 92.94\% with imagery content~\cite{huang2019identifying}. The study in \cite{feng2018extraction} explores different classical machine learning algorithms to classify relevant vs.\ irrelevant tweets using textual and imagery information. On the imagery content, they achieved an F1-score of 87.74\% using XGboost \cite{chen2016xgboost}.
The study in \cite{10.1145/3297280.3297481} proposes a simple, computationally inexpensive, multimodal two-stage framework to classify tweets (text and image) with built-infrastructure damage vs.\ nature-damage. The study investigated their approach using a home-grown dataset, and the SUN dataset \cite{5539970}. 
The study by Mouzannar et al.~\cite{Mouzannar2018} proposed a multimodal dataset, 
which has been developed for training a damage detection model. Similarly, Ofli et al. \cite{multimodalbaseline2020} explores unimodal as well as different multimodal modeling approaches based on a collection of multimodal social media posts. 

\subsection{Transfer Learning for Image Classification}
For the image classification task, transfer learning has been a popular approach, where a pre-trained neural network is used to train a model for a new task~\cite{yosinski2014transferable,sharif2014cnn,ozbulak2016transferable,oquab2014learning,multimodalbaseline2020,Mouzannar2018}. For this study, we follow the same approach using different deep learning architectures.

\subsection{Datasets}
Currently, publicly available datasets include damage severity assessment dataset~\cite{nguyen17damage}, CrisisMMD~\cite{alam2018crisismmd} and damage identification multimodal dataset~\cite{Mouzannar2018}. The first dataset is only annotated for images, whereas the last two are annotated for both text and images. Other relevant datasets are Disaster Image Retrieval from Social Media (DIRSM) \cite{bischke2017multimedia} and MediaEval 2018 \cite{alex258247}. 
The dataset reported in \cite{Gupta_2019_CVPR_Workshops} is constructed for detecting damage as an anomaly using pre-and post-disaster images. It consists of 700,000 building annotations.
A similar and relevant work is the development of the Incidents dataset \cite{weber2020detecting}, which consists of 446684 manually labeled Web images with 43 incident categories. The \textit{Crisis Benchmark Dataset} reported in \cite{FAlam:ASONAM20} is the largest so far for social media disaster image classification. 

For this study, we use the \textit{Crisis Benchmark Dataset}, and our study differs from \cite{FAlam:ASONAM20} in a number of ways. We provide more detailed experimental results on dataset comparison (i.e., individual vs.\ consolidated), compare different network architectures with a statistical significance test, and report the efficacy of data augmentation. We have also utilized a large unlabeled dataset to enhance the capability of the current model. We created multitask data splits from \textit{Crisis Benchmark Dataset} and report experimental results using both missing/incomplete and complete labels, which can serve as a baseline for future works.

\section{Tasks and Datasets}
\label{sec:tasks}

For this study, we addressed four different disaster-related tasks that are important for humanitarian aid. Below we provide details of each task and the associated class labels.  

\subsection{Tasks}
\label{ssec:tasks}

\subsubsection{Disaster type detection}
\label{sssec:task_disaster_types}
When ingesting images from unfiltered social media streams, it is important to detect different disaster types automatically from these images. For instance, an image can depict a wildfire, flood, earthquake, hurricane, and other types of disasters. In the literature, disaster types have been defined in different hierarchical categories such as natural, human-induced, and hybrid \cite{shaluf2007disaster}. Natural disasters are events that result from natural phenomena (e.g., fire, flood, earthquake). Human-induced disasters result from human actions (e.g., terrorist attacks, accidents, wars, and conflicts). Hybrid disasters result from human actions, which affect natural phenomena afterward (e.g., deforestation results in soil erosion and climate change). The class labels for disaster type include {\em (i)} earthquake, {\em (ii)} fire, {\em (iii)} flood, {\em (iv)} hurricane, {\em (v)} landslide, {\em (vi)} other disaster--to cover all other disaster types (e.g., plane crash), and {\em (vii)}  not disaster--for images that do not show any identifiable disasters.

\subsubsection{Informativeness}
\label{sssec:task_informativeness}
Images posted on social media during disasters do not always contain informative (e.g., an image showing damaged infrastructure due to flood, fire, or any other disaster events) or useful content for humanitarian aid. It is necessary to remove any irrelevant or redundant content to facilitate crisis responders' efforts more effectively. Therefore, the purpose of this classification task is to filter out irrelevant images. The class labels for this task are {\em (i)} informative and {\em (ii)} not informative.

\subsubsection{Humanitarian}
\label{sssec:task_humanitarian}
An important aspect of crisis responders is to assist people based on their needs, which requires information to be classified into more fine-grained categories to take specific actions. In the literature, humanitarian categories often include \textit{affected individuals}; \textit{injured or dead people}; \textit{infrastructure and utility damage}; \textit{missing or found people}; \textit{rescue, volunteering, or donation effort}; and \textit{vehicle damage}~\cite{alam2018crisismmd}. In this study, we focus on four categories that are deemed to be the most prominent and important for crisis responders such as {\em (i)} affected, injured, or dead people, {\em (ii)} infrastructure and utility damage, {\em (iii)} rescue volunteering or donation effort, and {\em (iv)} not humanitarian. 

\subsubsection{Damage severity}
\label{ssec:task_damage_severity}
Assessing the severity of the damage is important to help the affected community during disaster events. The severity of damage can be assessed based on the physical destruction to a built structure visible in an image (e.g., destruction of bridges, roads, buildings, burned houses, and forests). Following the work reported in~\cite{nguyen17damage}, we define the categories for this classification task as {\em (i)} severe damage, {\em (ii)} mild damage, and {\em (iii)} little or none. 

\begin{figure}[t]
\centering
\includegraphics[width=0.7\textwidth]{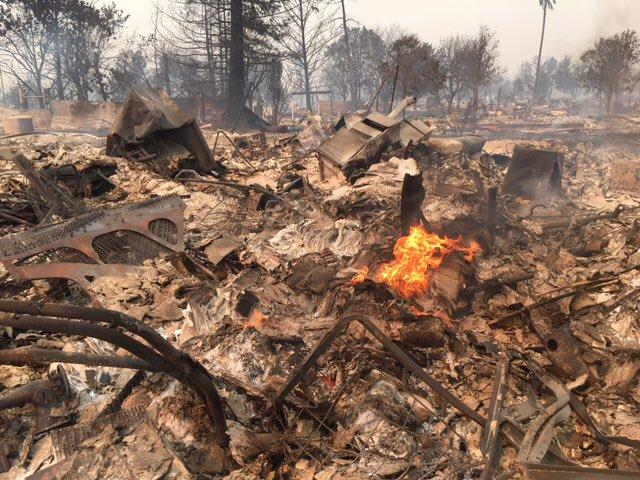}
\caption{An image annotated as {\em (i)}~fire event, {\em (ii)}~informative, {\em (iii)}~infrastructure and utility damage, and {\em (iv)}~severe damage.}
\label{fig:example_all_task}
\end{figure}

Figure \ref{fig:example_all_task} shows an example image with the labels for all four tasks. 

\subsection{Datasets}
\label{ssec:dataset}
As mentioned earlier, we used the dataset reported in \cite{FAlam:ASONAM20}.\footnote{\url{https://crisisnlp.qcri.org/crisis-image-datasets-asonam20}}  This dataset has been developed by curating existing publicly available sources, creating non-overlapping training, development, and test splits. For the sake of clarity and completeness, we provide a brief overview of the dataset. More details of the dataset curation and consolidation process can be found in \cite{FAlam:ASONAM20}. 

\subsubsection{Damage Assessment Dataset (DAD)}
\label{sssec:dataset_dad}
The damage assessment dataset consists of labeled imagery data with damage severity levels such as severe, mild, and little-to-no damage~\cite{nguyen17damage}. The images have been collected from two sources: AIDR~\cite{imran2014aidr} and Google. To crawl data from Google, authors used the following keywords: \textit{damage building, damage bridge, and damage road}. The images from AIDR were collected from Twitter during different disaster events such as Typhoon Ruby, Nepal Earthquake, Ecuador Earthquake, and Hurricane Matthew. The dataset contains $\sim$25K images annotated by paid workers as well as volunteers. In this study, we use this dataset for the informativeness and damage severity tasks. For the informativeness task, the study in \cite{FAlam:ASONAM20} mapped the \textit{mild} and \textit{severe} images into informative class and manually categorized the \textit{little-to-no damage} images into \textit{informative} and \textit{not informative} categories. For the damage severity task, the label \textit{little-to-no damage} mapped into \textit{little or none} to align with other datasets.

\subsubsection{CrisisMMD}
\label{sssec:dataset_crisismmd}
This is a multimodal (i.e., text and image) dataset, which consists of 18,082 images collected from tweets during seven disaster events crawled by the AIDR system~\cite{alam2018crisismmd}. The data is annotated by crowd workers using the Figure-Eight platform\footnote{Currently acquired by \url{https://appen.com/}} for three different tasks: {\em (i)} informativeness with binary labels (i.e., informative vs. not informative), {\em (ii)} humanitarian with seven class labels (i.e., ``infrastructure and utility damage'', ``vehicle damage'', ``rescue, volunteering, or donation effort'', ``injured or dead people'', ``affected individuals'', ``missing or found people'', ``other relevant information'' and ``not relevant''), {\em (iii)} damage severity assessment with three labels (i.e., severe, mild and ``little or no damage''). 
For the humanitarian task similar class labels are grouped together. The images with labels \textit{injured or dead people} and \textit{affected individuals} are mapped into one class label \textit{affected, injured, or dead people}; \textit{infrastructure and utility damage} and \textit{vehicle damage} are mapped into \textit{infrastructure and utility damage};  \textit{other relevant information}, and \textit{not relevant} are mapped into \textit{not humanitarian}. The images with label \textit{missing or found people} are removed as it is difficult to identify. This results in four class labels for humanitarian task. 

\subsubsection{AIDR Disaster Type Dataset (AIDR-DT)}
\label{sssec:dataset_aidr_dt}
AIDR-DT dataset consists of tweets collected from 17 disaster events and 3 general collections. The tweets of these collections have been collected by the AIDR system \cite{imran2014aidr}. The 17 disaster events include flood, earthquake, fire, hurricane, terrorist attack, and armed-conflict. The tweets in general collections contain keywords related to natural disasters, human-induced disasters, and security incidents. Images are crawled from these collections for disaster type annotation.
The labeling of these images was performed in two steps. First, a set of images were labeled as \textit{earthquake}, \textit{fire}, \textit{flood}, \textit{hurricane}, and \textit{none of these categories}. Then, a sample of $\sim$2,200 images labeled as \textit{none of these categories} in the previous step are selected for annotating \textit{not disaster} and \textit{other disaster} categories. 

For the landslide category, images are crawled from Google, Bing, and Flickr using keywords landslide, mudslide, ``mud slides'', landslip, ``rock slides'', rockfall, ``land slide'', earthslip, rockslide, and ``land collapse''. As images have been collected from different sources, therefore, it resulted in having duplicates. Duplicate filtering has been applied to remove exact- and near-duplicate images to resolve this issue. Then, the remaining images were manually labeled as \textit{landslide} and \textit{not landslide}. The resulted annotated dataset consists of labeled images with seven categories defined in Section \ref{sssec:task_disaster_types}.

\subsubsection{Damage Multimodal Dataset (DMD)}
\label{sssec:dataset_dmd}
The multimodal damage identification dataset consists of 5,878 images collected from Instagram and Google~\cite{Mouzannar2018}. The authors of the study crawled the images using more than 100 hashtags, which are proposed in crisis lexicon \cite{olteanu2014crisislex}. The manually labeled data consist of six damage class labels: fires, floods, natural landscape, infrastructural, human, and non-damage. The non-damage image includes cartoons, advertisements, and images that are not relevant or useful for humanitarian tasks. The study by Alam et al.~\cite{FAlam:ASONAM20} re-labeled images for all four tasks: disaster type, informativeness, humanitarian, and damage severity using the same class labels discussed in the previous section.

\subsection{Data Analysis}
\label{ssec:data_analysis}
\begin{figure*}[!htb]
        \centering
        \begin{subfigure}{.4\textwidth}
            \includegraphics[width=\textwidth]{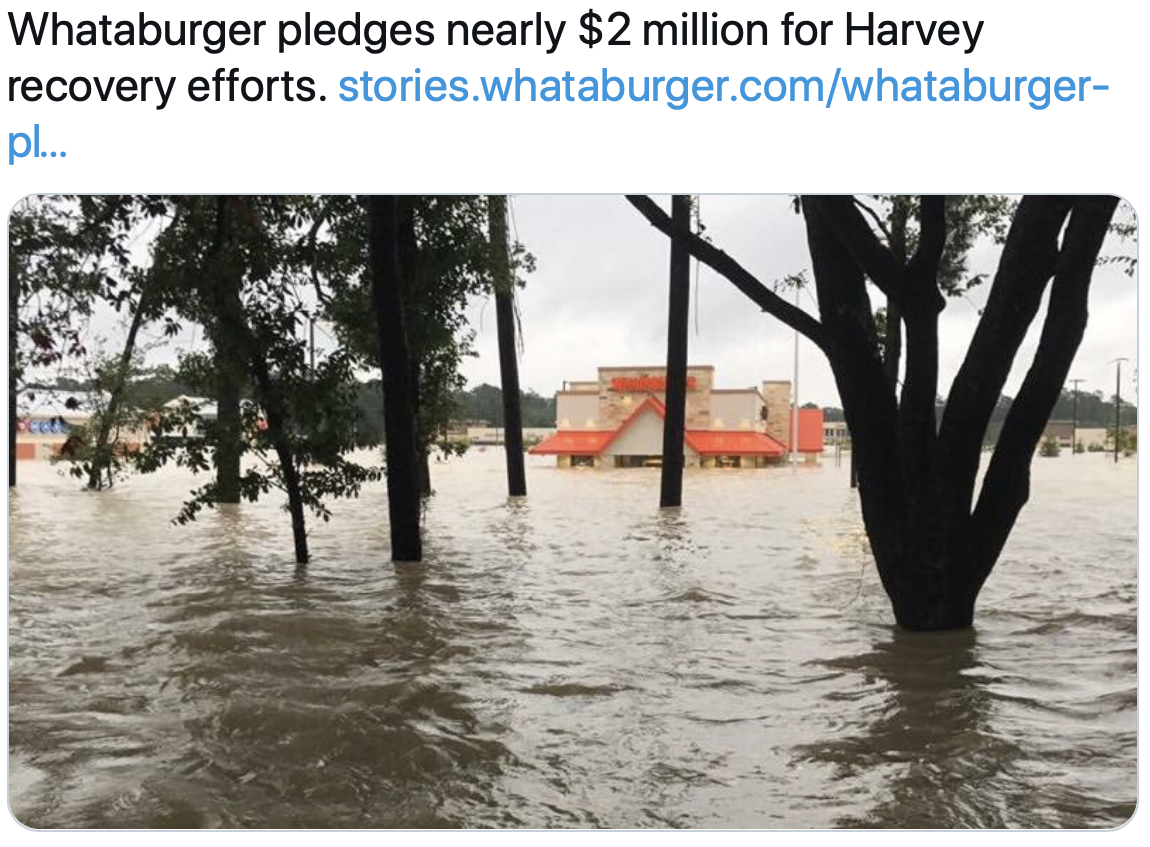}
            \caption{Image shared by a verified user.}
            \label{fig:image_shared_verified_user}
        \end{subfigure}\hspace{1em}%
        \begin{subfigure}{.4\textwidth}
                \includegraphics[width=\textwidth]{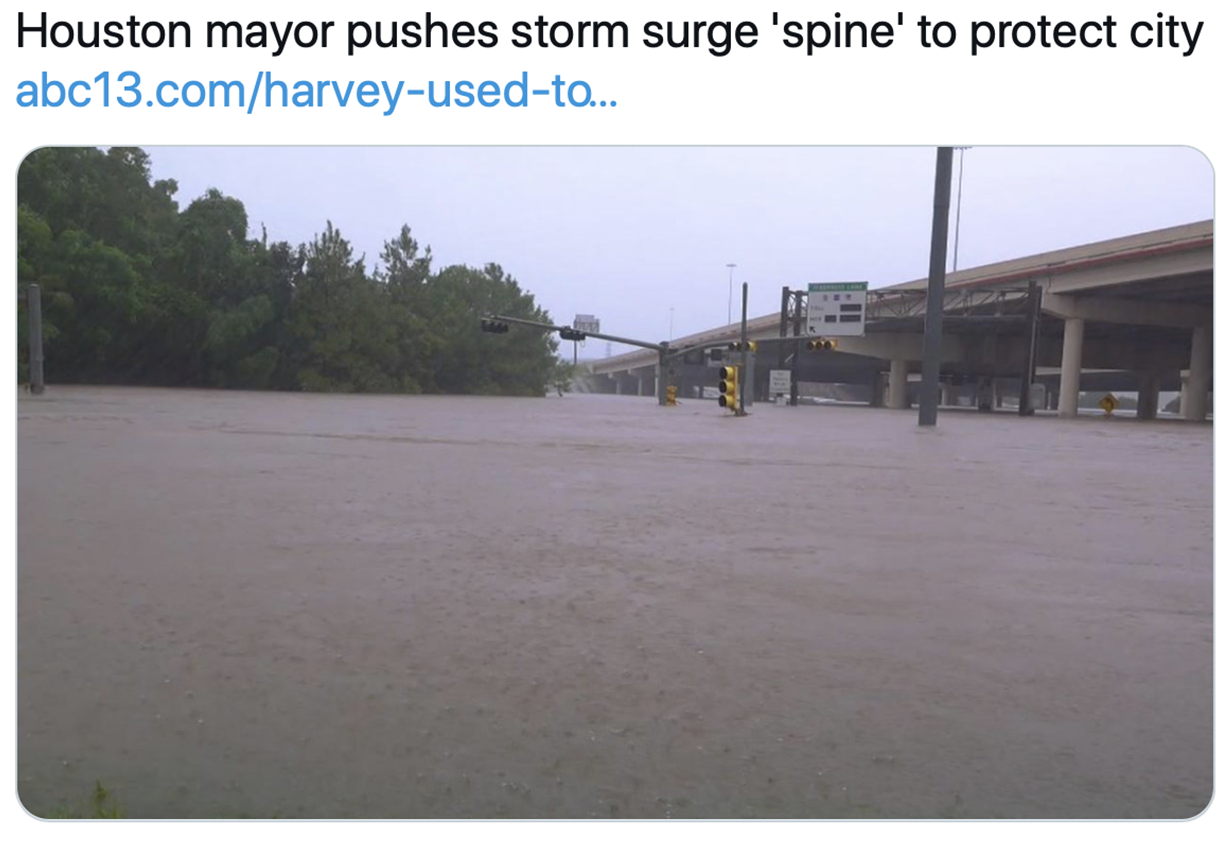}
                \caption{Image shared by an unverified user.}
                \label{fig:image_shared_unverified_user}
        \end{subfigure}
    \caption{Images shared by verified vs. unverified users. Both images show severe infrastructure damage.}
    \label{fig:image_sharing_behaviors}
\end{figure*}

To understand different aspects of the dataset, we analyze the distribution of images shared during different events, images shared by a different type of users (e.g., verified vs.\ unverified), and other characteristics. The dataset comprises images collected from different sources such as Google, Bing, Yahoo, and Twitter. Since only the images collected from Twitter contain social media information, we analyzed only those images that have Twitter's JSON objects ($\sim$27K images). In Table \ref{tab:tweet_image_dist}, we report statistics of the collected tweets and images for different events. It appears that people share images in only 1 to 5\% of the posts. We investigated the effect of the images shared by verified vs.\ unverified users. In Figure \ref{fig:image_sharing_behaviors}, we show two example images, one from a verified user (a) and another from an unverified user (b). We notice that images shared by verified users get more retweets than those shared by unverified users. For example, the image in Figure \ref{fig:image_shared_verified_user} has been retweeted 4,268 times and liked 11.7K times whereas Figure \ref{fig:image_shared_unverified_user} has not been retweeted even though it shows similar severe infrastructure damage. Among $\sim$27K images, there are 5,527 images with verified users and 22,207 images with unverified users. The users who shared a higher number of images are mostly news agencies. For example, in the annotated $\sim$27K images, we found that \textit{California Top News}\footnote{\url{https://twitter.com/CaliforniaBits}} shared 49 images during 2017 California wildfires, and among them 30 of the images are about \textit{``infrastructure and utility damage''} or \textit{``rescue volunteering or donation effort''}.

\begin{figure}[h!]
	\centering
	\includegraphics[width=0.70\linewidth]{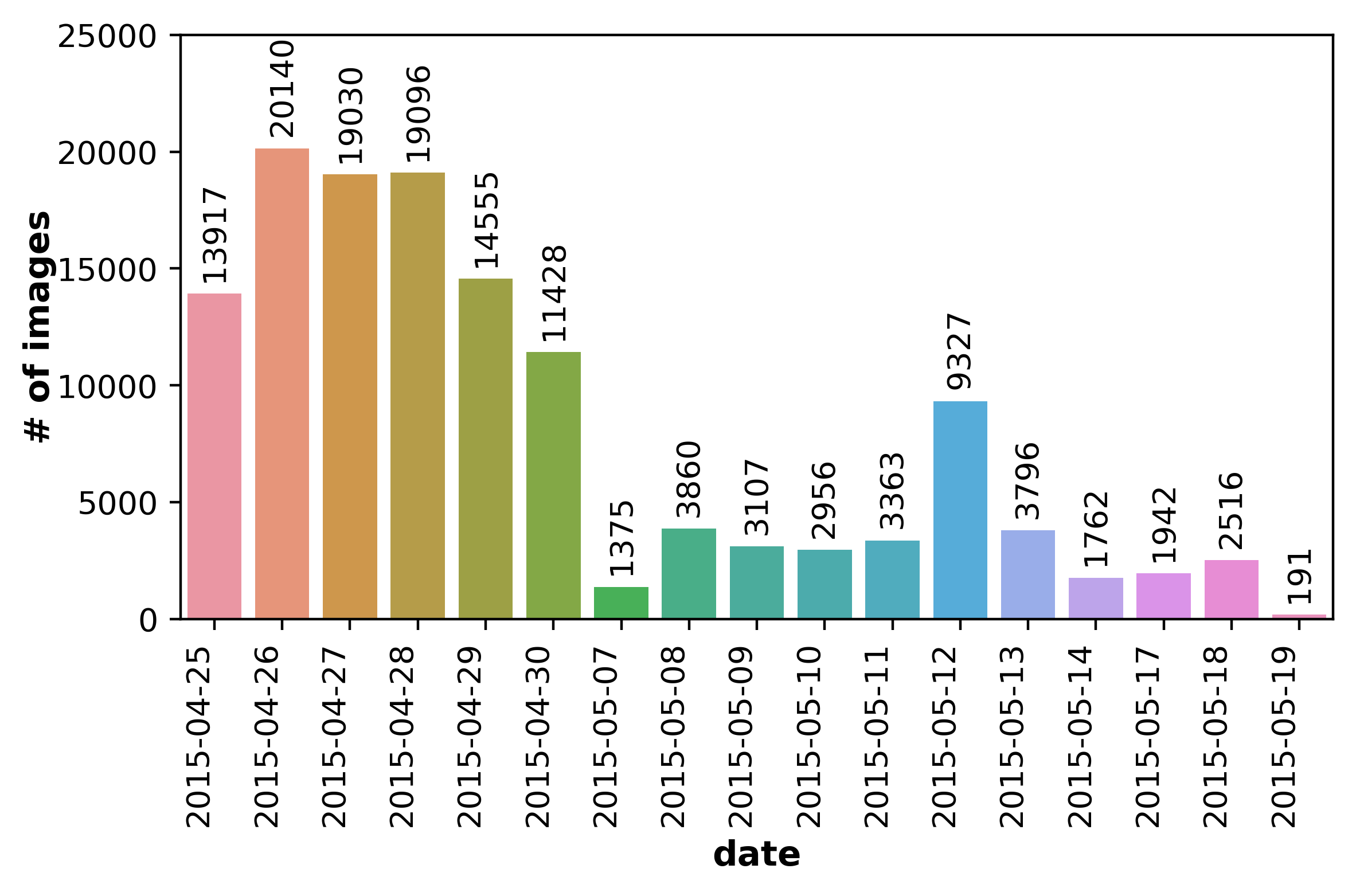}
	\caption{Number of images shared during 2015 Nepal Earthquake.}
	\label{fig:number_image_nepal_earthquake}
\end{figure}

We also analyzed image sharing behavior during disaster events where we considered all collected images with or without labels.\footnote{Note that only some of the collected images have been manually annotated.} We observed that more images are posted during the early days of a disaster and it gradually decreases, as illustrated in Figure \ref{fig:number_image_nepal_earthquake}.

\begin{table}[]
\centering
\setlength{\tabcolsep}{2.5pt}
\scalebox{0.8}{
\begin{tabular}{lrrrrrr}
\toprule
\multicolumn{1}{c}{\textbf{Event name}} & \multicolumn{1}{c}{\textbf{Year}} & \multicolumn{1}{c}{\textbf{\# tweets}} & \multicolumn{1}{c}{\textbf{\# images}} & \multicolumn{1}{c}{\textbf{\% of images}} & \multicolumn{1}{c}{\textbf{Start Date}} & \multicolumn{1}{c}{\textbf{End date}} \\ \midrule
Nepal earthquake & 2015 & 4,223,936 & 132,361 & 3.13 & 25-Apr-2015 & 19-May-2015 \\
Paris attack & 2015 & 10,599,629 & 499,953 & 4.72 & 14-Nov-2015 & 3-Dec-2015 \\
South india floods & 2015 & 2,994,119 & 141,831 & 4.74 & 3-Dec-2015 & 6-Dec-2015 \\
Food insecurity in Yemen & 2015 & 1,107,931 & 63,686 & 5.75 & 25-Sep-2015 & 19-Nov-2015 \\
Terremotoitalia & 2016 & 3,382,698 & 167,331 & 4.95 & 26-Oct-2016 & 27-Nov-2016 \\
Hurricane Irma & 2017 & 3,517,280 & 176,972 & 5.03 & 6-Sep-2017 & 21-Sep-2017 \\
Hurricane Harvey & 2017 & 6,664,349 & 321,435 & 4.82 & 26-Aug-2017 & 20-Sep-2017 \\
Hurricane Maria & 2017 & 2,953,322 & 52,231 & 1.77 & 20-Sep-2017 & 13-Nov-2017 \\
Mexico earthquake & 2017 & 383,341 & 7,111 & 1.86 & 20-Sep-2017 & 6-Oct-2017 \\
California wildfires & 2017 & 455,311 & 10,130 & 2.22 & 10-Oct-2017 & 27-Oct-2017 \\
Iraq-Iran earthquake & 2017 & 207,729 & 6,307 & 3.04 & 13-Nov-2017 & 19-Nov-2017 \\
Sri Lanka floods & 2017 & 41,809 & 2,108 & 5.04 & 31-May-2017 & 3-Jul-2017 \\
Syria attacks & 2017 & 5,381,866 & 107,513 & 2.00 & 6-Apr-2017 & 26-Apr-2017 \\
Ukraine conflict & 2017 & 1,268,942 & 30,289 & 2.39 & 5-Nov-2017 & 13-Nov-2017 \\
Kerala flood & 2018 & 3,044,703 & 15,767 & 0.52 & 17-Aug-2018 & 12-Sep-2018 \\
Hurricane Florence & 2018 & 623,074 & 12,879 & 2.07 & 11-Sep-2018 & 24-Sep-2018 \\
Hurricane Michael & 2018 & 243,263 & 5,106 & 2.10 & 10-Oct-2018 & 27-Oct-2018 \\ \bottomrule
\end{tabular}
}
\caption{Number of tweets and images collected during different disaster events.}
\label{tab:tweet_image_dist}
\end{table}

\subsection{Data Split}
Before consolidating the datasets, each dataset has been divided into training (train), development (dev), and test sets with 70:10:20 ratio, respectively. The purpose was threefold: {\em (i)}~train and evaluate individual datasets on each task, {\em (ii)}~have a close-to-equal distribution from each dataset into the final consolidated dataset, and {\em (iii)}~provide the research community an opportunity to use the splits independently. After data split, duplicate images are identified across sets and moved into the training set to create a non-overlapping test set.

\subsection{Data Consolidation}
\label{ssec:data_consolidation}
The primary motivation to perform data consolidation is to develop robust deep learning models with large amounts of data. For this purpose, all train, dev, and test sets are merged into the consolidated train, dev, and test sets, respectively. While doing so, duplicate images from the dev and test sets are moved into the train set to create non-overlapping splits. More detail about the duplicate identification process can be found in \cite{FAlam:ASONAM20}.  

\subsection{Data Statistics}
\label{ssec:class_label_dist}
Tables \ref{tab:data_split_disaster_types}, \ref{tab:data_split_informativemess}, \ref{tab:data_split_humanitarian}, \ref{tab:data_split_damage_severity}, and \ref{tab:consolidated_dataset} show the label distribution of all datasets for all four tasks. 
Some class labels are skewed in individual datasets. For example, in disaster type datasets (Table \ref{tab:data_split_disaster_types}), the distribution of the ``other disaster'' label is low in the AIDR-DT dataset, whereas the distribution of the ``landslide'' label low in the DMD dataset. For the informativeness task, low distribution is observed for the ``informative'' label. Moreover, for the humanitarian task, we have low distribution for the ``rescue volunteering or donation effort'' label in the DMD dataset, and for the damage severity task ``mild'' label in CrisisMMD and DMD datasets. However, the consolidated dataset creates a fair balance across class labels for different tasks, as shown in Table~\ref{tab:consolidated_dataset}.

\begin{table}[h]
\centering
\caption{Data split for the \textbf{disaster type} task. %
}
\label{tab:data_split_disaster_types}
\scalebox{0.9}{
\begin{tabular}{@{}llrrrr@{}}
\toprule
\multicolumn{1}{@{}l}{\textbf{Dataset}} & \multicolumn{1}{l}{\textbf{Class labels}} & \multicolumn{1}{r}{\textbf{Train}} & \multicolumn{1}{r}{\textbf{Dev}} & \multicolumn{1}{r}{\textbf{Test}} & \multicolumn{1}{r@{}}{\textbf{Total}} \\ \midrule
\multirow{8}{*}{\textbf{\begin{tabular}[c]{@{}l@{}}AIDR-DT \\
\end{tabular}}} & Earthquake & 1,910 & 201 & 376 & 2,487 \\
 & Fire & 990 & 105 & 214 & 1,309 \\
 & Flood & 2,059 & 241 & 533 & 2,833 \\
 & Hurricane & 1,188 & 142 & 279 & 1,609 \\
 & Landslide & 901 & 119 & 257 & 1,277 \\
 & Not disaster & 1,507 & 198 & 415 & 2,120 \\
 & Other disaster & 65 & 6 & 17 & 88 \\ \cmidrule(l){2-6}
 & \textbf{Total} & 8,620 & 1,012 & 2,091 & 11,723 \\ \midrule
\multirow{8}{*}{\textbf{\begin{tabular}[c]{@{}l@{}}DMD \\
\end{tabular}}} & Earthquake & 130 & 17 & 35 & 182 \\
 & Fire & 255 & 36 & 71 & 362 \\
 & Flood & 263 & 35 & 70 & 368 \\
 & Hurricane & 253 & 36 & 73 & 362 \\
 & Landslide & 38 & 5 & 11 & 54 \\
 & Not disaster & 2,108 & 288 & 575 & 2,971 \\
 & Other disaster & 1,057 & 145 & 287 & 1,489 \\ \cmidrule(l){2-6} 
 & \textbf{Total} & 4,152 & 506 & 1,130 & 5,788 \\ 
 \bottomrule %
\end{tabular}
}
\end{table}

\begin{table}[h]
\centering
\caption{Data split for the \textbf{informativeness} task.}
\label{tab:data_split_informativemess}
\scalebox{0.9}{
\begin{tabular}{@{}llrrrr@{}}
\toprule
\multicolumn{1}{@{}l}{\textbf{Dataset}} & \multicolumn{1}{l}{\textbf{Class labels}} & \multicolumn{1}{r}{\textbf{Train}} & \multicolumn{1}{r}{\textbf{Dev}} & \multicolumn{1}{r}{\textbf{Test}} & \multicolumn{1}{r@{}}{\textbf{Total}} \\ \midrule
\multirow{2}{*}{\textbf{\begin{tabular}[c]{@{}l@{}}DAD \\
\end{tabular}}} & Informative & 15,329 & 590 & 2,266 & 18,185 \\
 & Not informative & 5,950 & 426 & 1,259 & 7,635 \\ \cmidrule(l){2-6} 
 & \textbf{Total} & 21,279 & 1,016 & 3,525 & 25,820 \\ \bottomrule
\multirow{2}{*}{\textbf{\begin{tabular}[c]{@{}l@{}}CrisisMMD \\
\end{tabular}}} & Informative & 7,233 & 635 & 1,507 & 9,375 \\
 & Not informative & 6,535 & 551 & 1,621 & 8,707 \\ \cmidrule(l){2-6} 
 & \textbf{Total} & 13,768 & 1,186 & 3,128 & 18,082 \\ \bottomrule
\multirow{2}{*}{\textbf{\begin{tabular}[c]{@{}l@{}}DMD \\
\end{tabular}}} & Informative & 2,071 & 262 & 573 & 2,906 \\
 & Not informative & 2,152 & 240 & 580 & 2,972 \\ \cmidrule(l){2-6} 
 & \textbf{Total} & 4,223 & 502 & 1,153 & 5,878 \\ \bottomrule
\multirow{2}{*}{\textbf{\begin{tabular}[c]{@{}l@{}}AIDR-Info \\
\end{tabular}}} & Informative & 627 & 66 & 172 & 865 \\
 & Not informative & 6,677 & 598 & 1,796 & 9,071 \\ \cmidrule(l){2-6} 
 & \textbf{Total} & 7,304 & 664 & 1,968 & 9,936 \\ 
 \bottomrule
 \end{tabular}
}
\end{table}

\begin{table}[h]
\centering
\caption{Data split for the \textbf{humanitarian} task.}
\label{tab:data_split_humanitarian}
\scalebox{0.9}{
\begin{tabular}{@{}llrrrr@{}}
\toprule
\multicolumn{1}{@{}l}{\textbf{Dataset}} & \multicolumn{1}{l}{\textbf{Class labels}} & \multicolumn{1}{r}{\textbf{Train}} & \multicolumn{1}{r}{\textbf{Dev}} & \multicolumn{1}{r}{\textbf{Test}} & \multicolumn{1}{r@{}}{\textbf{Total}} \\ \midrule
\multirow{5}{*}{\textbf{\begin{tabular}[c]{@{}l@{}}CrisisMMD\\ %
\end{tabular}}} & Affected, injured, or dead people & 521 & 51 & 100 & 672 \\
 & Infrastructure and utility damage & 3,040 & 299 & 589 & 3,928 \\
 & Not humanitarian & 3,307 & 296 & 807 & 4,410 \\
 & Rescue volunteering or donation effort & 1,682 & 174 & 375 & 2,231 \\ \cmidrule(l){2-6} 
 & \textbf{Total}  & 8,550 & 820 & 1,871 & 11,241 \\ \midrule
\multirow{5}{*}{\textbf{\begin{tabular}[c]{@{}l@{}}DMD\\ %
\end{tabular}}} & Affected, injured, or dead people & 242 & 28 & 63 & 333 \\
 & Infrastructure and utility damage & 933 & 125 & 242 & 1,300 \\
 & Not humanitarian & 2,736 & 314 & 744 & 3,794 \\
 & Rescue volunteering or donation effort & 74 & 9 & 18 & 101 \\ \cmidrule(l){2-6} 
 & \textbf{Total} & 3,985 & 476 & 1,067 & 5,528 \\  
 \bottomrule
\end{tabular}
}
\end{table}

\begin{table}[h]
\centering
\caption{Data split for the \textbf{damage severity} task.}
\label{tab:data_split_damage_severity}
\scalebox{0.9}{
\begin{tabular}{@{}llrrrr@{}}
\toprule
\multicolumn{1}{@{}l}{\textbf{Dataset}} & \multicolumn{1}{l}{\textbf{Class labels}} & \multicolumn{1}{r}{\textbf{Train}} & \multicolumn{1}{r}{\textbf{Dev}} & \multicolumn{1}{r}{\textbf{Test}} & \multicolumn{1}{r@{}}{\textbf{Total}} \\ \midrule
\multirow{4}{*}{\textbf{\begin{tabular}[c]{@{}l@{}}DAD\\ %
\end{tabular}}} & Little or none & 7,881 & 1,101 & 1,566 & 10,548 \\
 & Mild & 2,828 & 388 & 546 & 3,762 \\
 & Severe & 9,457 & 673 & 1,380 & 11,510 \\ \cmidrule(l){2-6}
 & \textbf{Total} & 20,166 & 2,162 & 3,492 & 25,820 \\ \midrule
\multirow{4}{*}{\textbf{\begin{tabular}[c]{@{}l@{}}CrisisMMD\\ %
\end{tabular}}} & Little or none & 317 & 35 & 67 & 419 \\
 & Mild & 547 & 56 & 125 & 728 \\
 & Severe & 1,629 & 144 & 278 & 2,051 \\ \cmidrule(l){2-6}
 & \textbf{Total} & 2,493 & 235 & 470 & 3,198 \\\midrule
\multirow{4}{*}{\textbf{\begin{tabular}[c]{@{}l@{}}DMD \\ %
\end{tabular}}} & Little or none & 2,874 & 331 & 778 & 3,983 \\
 & Mild & 508 & 60 & 132 & 700 \\
 & Severe & 857 & 110 & 228 & 1,195 \\ \cmidrule(l){2-6}
 & \textbf{Total} & 4,239 & 501 & 1,138 & 5,878 \\  
 \bottomrule
\end{tabular}
}
\end{table}

\begin{table}[h]
\centering
\caption{Data splits for the \textbf{consolidated dataset} for all tasks.}
\label{tab:consolidated_dataset}
\scalebox{0.80}{
\begin{tabular}{@{}lrrrr@{}}
\toprule
\multicolumn{1}{@{}l}{\textbf{Class labels}} & \multicolumn{1}{r}{\textbf{Train}} & \multicolumn{1}{r}{\textbf{Dev}} & \multicolumn{1}{r}{\textbf{Test}} & \multicolumn{1}{r@{}}{\textbf{Total}} \\ \midrule
\multicolumn{5}{c}{\textbf{Disaster type}} \\ \midrule %
Earthquake & 2,058 & 207 & 404 & 2,669 \\
Fire & 1,270 & 121 & 280 & 1,671 \\
Flood & 2,336 & 266 & 599 & 3,201 \\
Hurricane & 1,444 & 175 & 352 & 1,971 \\
Landslide & 940 & 123 & 268 & 1,331 \\
Not disaster & 3,666 & 435 & 990 & 5,091 \\
Other disaster & 1,132 & 143 & 302 & 1,577 \\ \cmidrule(l){2-5}
\textbf{Total} & 12,846 & 1,470 & 3,195 & 17,511 \\ \midrule
\multicolumn{5}{c}{\textbf{Informativeness}} \\\midrule %
Informative & 26,486 & 1,432 & 3,414 & 31,332 \\
Not informative & 21,700 & 1,622 & 5,063 & 28,385 \\ \cmidrule(l){2-5}
\textbf{Total} & 48,186 & 3,054 & 8,477 & 59,717 \\ \midrule 
\multicolumn{5}{c}{\textbf{Humanitarian }} \\ \midrule %
Affected, injured, or dead people & 772 & 73 & 160 & 1,005 \\
Infrastructure and utility damage & 4,001 & 406 & 821 & 5,228 \\
Not humanitarian & 6,076 & 578 & 1,550 & 8,204 \\
Rescue volunteering or donation effort & 1,769 & 172 & 391 & 2,332 \\  \cmidrule(l){2-5}
\textbf{Total} & 12,618 & 1,229 & 2,922 & 16,769 \\ \midrule 
\multicolumn{5}{c}{\textbf{Damage severity }} \\\midrule %
Little or none & 11,437 & 1,378 & 2,135 & 14,950 \\
Mild & 4,072 & 489 & 629 & 5,190 \\
Severe & 12,810 & 845 & 1,101 & 14,756 \\ \cmidrule(l){2-5}
\textbf{Total} & 28,319 & 2,712 & 3,865 & 34,896 \\ 
\bottomrule
\end{tabular}
}
\end{table}

\section{Experiments}
\label{sec:experiments}

Our experiments consists of {\em(i)} individual vs.\ consolidated datasets comparison (\textit{RQ1}), {\em(ii)} neural network architectures comparison (\textit{RQ2}) on the consolidated dataset, {\em(iii)} data augmentation (\textit{RQ3}), {\em(iv)} semi-supervised learning (\textit{RQ3}), and {\em(iv)} multitask learning (\textit{RQ4}). Below we first provide experimental settings, and then, discuss different experiments that we conducted for this study.  

\subsection{Experimental Setup}
\label{ssec:exp_settings}
We employ the transfer learning approach to perform experiments, which has shown promising results for various visual recognition tasks in the literature~\cite{yosinski2014transferable,sharif2014cnn,ozbulak2016transferable,oquab2014learning}. The idea of the transfer learning approach is to use existing weights of a pre-trained model for different downstream tasks. For this study, we used several neural network architectures using the PyTorch library.\footnote{https://pytorch.org/} The architectures include ResNet18, ResNet50, ResNet101~\cite{he2016deep}, AlexNet~\cite{krizhevsky2012imagenet}, VGG16~\cite{simonyan2014very},  DenseNet~\cite{huang2017densely}, SqueezeNet~\cite{i2016squeezenet}, InceptionNet~\cite{szegedy2016rethinking}, MobileNet~\cite{howard2017mobilenets}, and EfficientNet~\cite{tan2019efficientnet}.  

We use the weights of the networks pre-trained using ImageNet~\cite{deng2009imagenet} to initialize our model. 
We adapt the last layer (i.e., softmax layer) of the network according to the particular classification task at hand instead of the original 1,000-way classification. The transfer learning approach allows us to transfer the features and the parameters of the network from the broad domain (i.e., large-scale image classification) to the specific one. Put specifically, we designed a binary classifier for the informativeness task and multi-class classifiers for the remaining three tasks.
We train the models using the Adam optimizer~\cite{kingma2014adam} %
with an initial learning rate of $10^{-5}$, which is decreased by a factor of 10 %
when accuracy on the dev set stops improving for 10 epochs. The models were trained for 150 epochs. 
To measure the performance of each classifier, we use weighted average precision (P), recall (R), and F1-score (F1).

\subsection{Dataset Comparison}
\label{ssec:exp_dataset_comp}
To determine whether consolidated data helps in achieving better performance, we train the models using training sets from the individual and consolidated datasets. However, we always test the models on the consolidated test set. As our test data is the same across different experiments, this ensures that results are comparable. Since we have four different tasks, consisting of fifteen different datasets, we only experimented with the ResNet18~\cite{he2016deep} network architecture to manage the computational load.

\subsection{Network Architectures}
\label{ssec:exp_net_arch}
Currently available neural network architectures come with different computational complexity. As one of our goals is to deploy the models in real-time applications, we exploit them to understand their performance differences. Another motivation is that current literature in crisis informatics only reports results using one or two network architectures (e.g., VGG16 in \cite{multimodalbaseline2020}, InceptionNet in \cite{Mouzannar2018}), which may lead to sub-optimal outcomes.

\subsection{Data Augmentation}
\label{ssec:exp_data_augmentation}
Data augmentation is a commonly used technique to improve the generalization of deep neural networks in the absence of large-scale datasets. We experiment with the recently proposed RandAugment \cite{cubuk2020randaugment} method for image augmentation. In literature, RandAugment was proposed as a fast alternative for learned augmentation strategies. We used the PyTorch implementation\footnote{\href{https://github.com/ildoonet/pytorch-randaugment}{https://github.com/ildoonet/pytorch-randaugment}} in our experiments. To increase the diversity of generated examples, we used the following 16 different transformations:

\setlength{\columnsep}{0.1in}
\begin{multicols}{4}
\begin{enumerate} %
 \itemsep0em
  \item AutoContrast
  \item Equalize
  \item Invert
  \item Rotate
  \item Color
  \item Posterize
  \item Solarize
  \item SolarizeAdd
  \item Contrast
  \item Brightness
  \item Sharpness
  \item ShearX
  \item ShearY
  \item CutoutAbs
  \item TranslateX
  \item TranslateY
\end{enumerate}
\end{multicols}

\noindent where augmentation strengths can be controlled with two tunable parameters: 
\begin{enumerate}
    \itemsep0em
    \item[] $N$: the number of augmentation transformations to apply sequentially
    \item[] $M$: magnitude for all the transformations.
\end{enumerate}

Each transformation resides on an integer scale from 0 to 30, with 30 being the maximum strength. In our experiments, we use constant magnitude $M$ for all augmentations. The augmentation method then boils down to randomly selecting $N$ transformations and applying each transformation sequentially with strength corresponding to scale $M$.  

In addition, we used \textit{weight decay}, which is one of the most commonly used techniques for regularizing parametric machine learning models \cite{moody1995simple}. This helps to reduce the overfitting of the models and avoids exploding gradient. 

We have conducted the data augmentation experiments using all ten different neural network architectures. We used a weight decay of $10^{-3}$ and other hyper-parameters remain the same as discussed in Section \ref{ssec:exp_settings}.

\subsection{Semi-supervised Learning}
\label{ssec:exp_semisupervised}
State-of-the-art image classification models are often trained with a large amount of labeled data, which is prohibitively expensive to collect in many applications. Semi-supervised learning is a powerful approach to mitigate this issue and leverage unlabeled data to improve the performance of machine learning models. Since unlabeled data can be obtained without significant human labor, performance boost gained from semi-supervised learning comes at low cost and can be scaled easily. In literature many semi-supervised techniques has been proposed focusing on deep learning \cite{xie2020self,sohn2020fixmatch,berthelot2019remixmatch,berthelot2019mixmatch,laine2016temporal,lee2013pseudo,mclachlan1975iterative,sajjadi2016regularization,tarvainen2017mean,verma2019interpolation,xie2019unsupervised,alam2018graph}. Among them self-training approach is one of the earliest \cite{scudder1965probability}, which has been adopted for deep neural network. The self-training approach, also called pseudo-labeling \cite{lee2013pseudo}, uses the model's prediction as a label and retrains the model against it.

For this study, we use \textit{Noisy student} (i.e., a simple self-training approach) training, which was proposed in \cite{xie2020self} as a semi-supervised learning approach to improve the accuracy and robustness of state-of-the-art image classification models. The algorithm consists of three main steps:
\begin{enumerate}[label={\textbf{Step \protect\threedigits{\theenumi}:}},leftmargin=*]
    \itemsep0em
    \item Train a teacher model on labeled images
    \item Use the teacher model to generate pseudo labels on unlabeled images
    \item Train a student model on combined labeled and pseudo labeled images
\end{enumerate}

The algorithm can be iterated multiple times by treating the student as the new teacher and labeling the unlabeled images with this model. During the learning phase of the student, different noises can be injected, such as dropout \cite{srivastava2014dropout} and data augmentation via RandAugment \cite{cubuk2020randaugment}. The student model is made larger than or equal to the teacher. The presence of noise and larger model capacity help the student model generalize better than the teacher.

\paragraph{Labeled dataset:}
As for the labeled dataset, we used our consolidated datasets and ran the experiments for all tasks. 

\paragraph{Unlabeled dataset:}
To obtain unlabeled images, we crawled images from the tweets of 20 different disaster collections (as mentioned in Section \ref{sssec:dataset_aidr_dt}). We removed duplicates and ensured the same images are not in our labeled dataset by matching their ids and applying duplicate filtering. The resulting unlabeled dataset consists of 1,514,497 images. 

\paragraph{Architecture:}  
We ran our experiments using the EfficientNet (b1) architecture as it performed better than the other models. In addition, it is one of the models used with \textit{Noisy student} experiments reported in \cite{xie2020self}. One significant difference between \cite{xie2020self} and our work is that we initialize our student model's weight with ImageNet pre-trained weights. In contrast, in \cite{xie2020self}, they train weights from scratch. Since our labeled dataset is significantly smaller than the ImageNet dataset, training from scratch substantially degrades performance in our experiments.

\paragraph{Training details:}
We first trained the model using the EfficientNet (b1) architecture on the labeled dataset (\textbf{Step 1}), which is referred to as the teacher model. We then predicted output for the unlabeled images (\textbf{Step 2}). After that, we trained the student EfficientNet (b1) model by combining labeled and pseudo-labeled images (\textbf{Step 3}). In this step, for the unlabeled data, we performed different filtering and balancing. We selected the images that have a confidence label greater than a certain task-specific threshold. After this, we balanced the training data so that each class has the same number of images as the class having the lowest number of images. To do this, for each class, we take the images having the highest confidence scores.

For the experiments, we used a batch size of 16 for labeled images and 48 for unlabeled images. Labeled and unlabeled images are concatenated together to compute the average cross-entropy loss. We used RandAugment with the number of augmentation, $N=5$, and the strength of augmentation, $M=12$.  We optimized the confidence thresholds separately for different tasks using the dev sets. The thresholds for disaster types, informativeness, humanitarian, and damage severity tasks were respectively 0.7, 0.8, 0.45, and 0.45. Similar to the data augmentation experiments, we used a weight decay of $10^{-3}$ and kept other hyper-parameters the same as discussed in Section \ref{ssec:exp_settings}.

\begin{table}[h]
\centering
\caption{Data split for multi-task setting with \textbf{incomplete/missing labels}. DS: Disaster types, Info: Informative, Hum: Humanitarian, DS: Damage Severity}
\label{tab:multitask-dataset}
\scalebox{0.88}{
\begin{tabular}{@{}lrrrr@{}}
\toprule
\multicolumn{1}{c}{\textbf{Class labels}} & \multicolumn{1}{c}{\textbf{Train}} & \multicolumn{1}{c}{\textbf{Dev}} & \multicolumn{1}{c}{\textbf{Test}} & \multicolumn{1}{c}{\textbf{Total}} \\ \midrule
\multicolumn{5}{c}{\textbf{Disaster types}} \\\midrule
Earthquake & 1,987 & 218 & 464 & 2,669 \\
Fire & 1,115 & 154 & 402 & 1,671 \\
Flood & 2,175 & 300 & 726 & 3,201 \\
Hurricane & 1,249 & 216 & 506 & 1,971 \\
Landslide & 917 & 127 & 287 & 1,331 \\
Not disaster & 3,064 & 564 & 1,463 & 5,091 \\
Other disaster & 489 & 218 & 870 & 1,577 \\\cmidrule{2-5}
\textbf{Total} & 10,996 & 1,797 & 4,718 & 17,511 \\ \midrule
\multicolumn{5}{c}{\textbf{Informative}} \\\midrule
Informative & 22,018 & 2,736 & 6,578 & 31,332 \\
Not informative & 18,841 & 2,460 & 7,084 & 28,385 \\\cmidrule{2-5}
\textbf{Total} & 40,859 & 5,196 & 13,662 & 59,717 \\\midrule
\multicolumn{5}{c}{\textbf{Humanitarian}} \\\midrule
Affected injured or dead people & 537 & 115 & 353 & 1,005 \\
Infrastructure and utility damage & 2,397 & 736 & 2,095 & 5,228 \\
Not humanitarian & 4,354 & 886 & 2,964 & 8,204 \\
Rescue volunteering or donation effort & 1,312 & 268 & 752 & 2,332 \\\cmidrule{2-5}
\textbf{Total} & 8,600 & 2,005 & 6,164 & 16,769 \\\midrule
\multicolumn{5}{c}{\textbf{Damage Severity}} \\\midrule
Little or none & 9,124 & 1,677 & 4,149 & 14,950 \\
Mild & 3,188 & 663 & 1,339 & 5,190 \\
Severe & 11,102 & 1,145 & 2,509 & 14,756 \\\cmidrule{2-5}
\textbf{Total} & 23,414 & 3,485 & 7,997 & 34,896 \\ \bottomrule
\end{tabular}
}
\end{table}

\begin{table}[h]
\centering
\caption{Data split for multitask setting with \textbf{complete aligned labels} for the different combinations of two-tasks.}
\label{tab:multitask-dataset_two_tasks}
\scalebox{0.88}{
\begin{tabular}{@{}lrrrr@{}}
\toprule
\multicolumn{5}{c}{\textbf{\textbf{Two tasks: Info and Hum}}} \\ \midrule
\multicolumn{1}{c}{\textbf{Class labels}} & \multicolumn{1}{c}{\textbf{Train}} & \multicolumn{1}{c}{\textbf{Dev}} & \multicolumn{1}{c}{\textbf{Test}} & \multicolumn{1}{c}{\textbf{Total}} \\ \midrule
\multicolumn{5}{c}{\textbf{Informative}} \\\midrule
Informative & 2,111 & 399 & 1,064 & 3,574 \\
Not informative & 2,546 & 397 & 1,443 & 4,386 \\ \cmidrule{2-5}
\textbf{Total} & 4,657 & 796 & 2,507 & 7,960 \\\midrule
\multicolumn{5}{c}{\textbf{Humanitarian}} \\\midrule
Affected injured or dead people & 426 & 72 & 166 & 664 \\
Infrastructure and utility damage & 410 & 81 & 210 & 701 \\
Rescue volunteering or donation effort & 1,274 & 246 & 688 & 2,208 \\
Not humanitarian & 2,547 & 397 & 1,443 & 4,387 \\ \cmidrule{2-5}
\textbf{Total} & 4,657 & 796 & 2,507 & 7,960 \\\midrule \midrule
\multicolumn{5}{c}{\textbf{Two tasks: Info and damage severity}} \\\midrule
\multicolumn{5}{c}{\textbf{Informative}} \\\midrule
Informative & 14,683 & 1,306 & 2,206 & 18,195 \\
Not informative & 4,687 & 928 & 2,020 & 7,635 \\ \cmidrule{2-5}
\textbf{Total} & 19,370 & 2,234 & 4,226 & 25,830 \\\midrule
\multicolumn{5}{c}{\textbf{Damage Severity}} \\\midrule
Little or none & 7,085 & 1,094 & 2,369 & 10,548 \\
Mild & 2,665 & 426 & 679 & 3,770 \\
Severe & 9,620 & 714 & 1,178 & 11,512 \\ \cmidrule{2-5}
\textbf{Total} & 19,370 & 2,234 & 4,226 & 25,830 \\ \bottomrule
\end{tabular}
}
\end{table}

\begin{table}[h]
\centering
\caption{Data split for multi-task setting with \textbf{complete aligned labels} for four-tasks: DS, Info, Hum and DS.}
\label{tab:multitask-dataset_four_tasks}
\scalebox{0.88}{
\begin{tabular}{@{}lrrrr@{}}
\toprule
\multicolumn{1}{c}{\textbf{Class labels}} & \multicolumn{1}{c}{\textbf{Train}} & \multicolumn{1}{c}{\textbf{Dev}} & \multicolumn{1}{c}{\textbf{Test}} & \multicolumn{1}{c}{\textbf{Total}} \\\midrule
\multicolumn{5}{c}{\textbf{Disaster types}} \\\midrule
Earthquake & 68 & 25 & 90 & 183 \\
Fire & 80 & 35 & 155 & 270 \\
Flood & 102 & 54 & 162 & 318 \\
Hurricane & 110 & 75 & 214 & 399 \\
Landslide & 8 & 6 & 24 & 38 \\
Other disaster & 372 & 198 & 806 & 1,376 \\ 
Not disaster & 1,563 & 368 & 1,043 & 2,974 \\ \cmidrule{2-5}
\textbf{Total} & 2,303 & 761 & 2,494 & 5,558 \\\midrule
\multicolumn{5}{c}{\textbf{Informative}} \\\midrule
Informative & 740 & 393 & 1,454 & 2,587 \\
Not informative & 1,563 & 368 & 1,040 & 2,971 \\ \cmidrule{2-5}
\textbf{Total} & 2,303 & 761 & 2,494 & 5,558 \\\midrule
\multicolumn{5}{c}{\textbf{Humanitarian}} \\\midrule
Affected injured or dead people & 85 & 34 & 164 & 283 \\
Infrastructure and utility damage & 398 & 230 & 764 & 1,392 \\
Rescue volunteering or donation effort & 26 & 14 & 53 & 93 \\ 
Not humanitarian & 1,794 & 483 & 1,513 & 3,790 \\ \cmidrule{2-5}
\textbf{Total} & 2,303 & 761 & 2,494 & 5,558 \\\midrule
\multicolumn{5}{c}{\textbf{Damage Severity}} \\\midrule
Little or none & 1,805 & 494 & 1,571 & 3,870 \\
Mild & 174 & 102 & 337 & 613 \\
Severe & 324 & 165 & 586 & 1075 \\ \cmidrule{2-5}
\textbf{Total} & 2,303 & 761 & 2,494 & 5,558 \\ \bottomrule
\end{tabular}
}
\end{table}

\subsection{Multi-task Learning}
\label{ssec:exp_multitask_learning}

Since the tasks share similar properties, we also consider training the model in multitask settings with shared parameters. The benefits of multitask settings can be twofold: {\em (i)} learning shared representation can help the model generalize better and improve performance on individual tasks, and {\em (ii)} training a single model instead of four different models will yield a significant speed and reduce computational load during training and inference. It is important to mention that the \textit{Crisis Benchmark Dataset} was not designed for multitask learning; rather, it was prepared for each task separately. Hence, we needed to prepare them for the multitask setup. Creating multitask learning datasets from \textit{Crisis Benchmark Dataset} introduced a challenge -- there is an overlap between train and test set images among different tasks. Hence, we prepare the datasets for the multitask setting using the following strategy:
\begin{enumerate}
    \itemsep0em
    \item We merge the test sets from different tasks into a combined test set. If an image in the combined test set is present in the train or dev set of some tasks, we remove it from that split and add the label of the task in the test set.
    \item We merge the dev sets of the four tasks into the combined dev set. If an image in the combined dev set is present in the train set of some tasks, we remove it from that train split and add the label of the task in the dev set. 
    \item We merge the train sets of the four tasks into the combined train set. Since we have removed images that overlap with the dev set and test set in the previous steps, therefore, it guarantees that no image from the train set will be present in the other splits.
\end{enumerate}

Since all the images do not have annotation for all four tasks, there is a discrepancy in the number of images available for different tasks. We report the distribution of the data splits for the multi-task setting in Table \ref{tab:multitask-dataset}. Overall, there are 49353 images in the train set, 6157 images in the dev set, and 15688 images in the test set. Due to the overlap of images in different splits for different tasks, there is also a discrepancy between the number of images available between multi-task and single-task settings. As an example, for the disaster types task, there are 12846 images in the train set, 1470 images in the dev set, and 3195 images in the test set in the single-task setting. However, in the multi-task setting, these numbers are respectively 10996, 1797, and 4718. As a consequence of our merging procedure, there are more images in the test and dev sets and fewer images in the train set.

Few approaches have been proposed in the literature to address the issue of incomplete/missing labels in multi-task settings. They usually work by generating missing task labels using different methods, including Bayesian networks \cite{kapoor2012multilabel}, rule-based approach \cite{kollias2019expression}, knowledge distillation from another model \cite{deng2020multitask}. In our experiments, we opt for a simpler alternative. Specifically, we do not compute loss for a task if its label is missing. Since the tasks have varying training images, we calculate the loss for each task and aggregate them in a batch. This ensures that the loss of each task is weighted equally. The steps are detailed in Algorithm \ref{algo-multitask}.

\newlength\mylen
\newcommand\myinput[1]{%
  \settowidth\mylen{\KwIn{}}%
  \setlength\hangindent{\mylen}%
  \hspace*{\mylen}#1}

\SetKwInput{KwInput}{Input}                %
\SetKwInput{KwOutput}{Output}              %

\begin{algorithm}
\caption{Batch loss calculation in the multi-task setting}
\label{algo-multitask}
\DontPrintSemicolon
\KwIn{batch\_input \tcp*{images in the batch}}
\myinput{batch\_labels\tcp*{list of labels for each task}} 
\myinput{num\_classes \tcp*{number of classes for each task}}
\myinput{model \tcp*{outputs prediction for all tasks are combined}} 
\KwOutput{batch\_loss}

num\_tasks = $len$(num\_classes) \;
prediction = $model.predict$(batch\_input) \;
batch\_loss = 0 \;
task\_index = 0 \tcp*{starting index for output corresponding to this task}

\For{$i\gets0$ \KwTo $num\_tasks$}{
    prediction\_task = prediction[:, task\_index:task\_index + num\_classes[i]] \;
    label\_task = batch\_labels[i] \;
    \tcc{if there is no label for a task it is marked as -1 in the label}
    valid\_idx = $nonzero$(label\_task != -1) \; 
    task\_loss = $cross\_entropy\_loss$(prediction\_task[valid\_idx], label\_task[valid\_idx]) \;
    batch\_loss = batch\_loss + task\_loss \;
    task\_index  = task\_index + num\_classes[i]
}
\end{algorithm}

We also experiment with images having complete aligned labels for different tasks. We identified three such combinations that have a substantial number of images in different classes. Two of them belong to two task subsets. The first one is informativeness and humanitarian, which has 7,960 total aligned images. The second one is informativeness and damage severity, having 25,830 total images. Data distribution for these two settings is reported in Table \ref{tab:multitask-dataset_two_tasks}. The final subset of images having labels for all four tasks, which consists of 5558 images. Data distribution for this set is reported in Table \ref{tab:multitask-dataset_four_tasks}.
\section{Results}
\label{sec:results}
Our experimental results consist of different settings. Below we discuss each of them in detail. 
\textit{\begin{table}[t]
\centering
\caption{Results on different classification tasks using the ResNet18 model. Trained on individual and consolidated datasets and tested on consolidated test sets.}
\label{tab:classification_results}
\scalebox{0.9}{
\begin{tabular}{@{}lrrrr@{}}
\toprule
\multicolumn{1}{@{}l}{\textbf{Dataset}} & \multicolumn{1}{c}{\textbf{Acc}} & \multicolumn{1}{c}{\textbf{P}} & \multicolumn{1}{c}{\textbf{R}} & \multicolumn{1}{c}{\textbf{F1}} \\ \midrule
\multicolumn{5}{c}{\textbf{Disaster type (7 classes)}} \\ \midrule
AIDR-DT & 0.76 & 0.72 & 0.76 & 0.73 \\
DMD & 0.58 & 0.73 & 0.58 & 0.59 \\
\rowcolor[HTML]{add8e6}
\textbf{Consolidated} & 0.79 & 0.78 & 0.79 & \underline{\textbf{0.79}} \\\midrule
\multicolumn{5}{c}{\textbf{Informativeness (2 classes)}} \\  \midrule
DAD & 0.80 & 0.80 & 0.80 & 0.80 \\
CrisisMMD & 0.79 & 0.79 & 0.79 & 0.79 \\
DMD & 0.80 & 0.80 & 0.80 & 0.80 \\
AIDR-Info & 0.75 & 0.79 & 0.75 & 0.73 \\
\rowcolor[HTML]{add8e6}
\textbf{Consolidated} & 0.85 & 0.85 & 0.85 & \underline{\textbf{0.85}} \\\midrule
\multicolumn{5}{c}{\textbf{Humanitarian (4 classes)}} \\ \midrule
CrisisMMD & 0.73 & 0.73 & 0.73 & 0.73 \\
DMD & 0.68 & 0.68 & 0.68 & 0.64 \\
\rowcolor[HTML]{add8e6}
\textbf{Consolidated} & 0.75 & 0.75 & 0.75 & \underline{\textbf{0.75}} \\\midrule
\multicolumn{5}{c}{\textbf{Damage severity (3 classes)}} \\ \midrule
DAD & 0.72 & 0.70 & 0.72 & 0.71 \\
CrisisMMD & 0.41 & 0.57 & 0.41 & 0.37 \\
DMD & 0.68 & 0.66 & 0.68 & 0.66 \\
\rowcolor[HTML]{add8e6}
\textbf{Consolidated} & 0.75 & 0.73 & 0.75 & \underline{\textbf{0.74}} \\ \bottomrule
\end{tabular}
}
\end{table}}

\subsection{Dataset Comparison}
\label{ssec:baseline_results_datasets}
In Table~\ref{tab:classification_results}, we report classification results for different tasks and different datasets using ResNet18 network architecture. The performance of different tasks are not equally comparable as they have different levels of complexity (e.g., varying number of class labels, class imbalance, etc.). For example, the informativeness classification is a binary task, which is computationally simpler than a classification task with more labels (e.g., seven labels in disaster type). Hence, the performance is comparatively higher for informativeness. An example of a class imbalance issue can be seen in Table \ref{tab:consolidated_dataset} with the damage severity task. The distribution of mild is relatively small, which reflects on its and overall performance. The mild class label is also less distinctive than other class labels, and we noticed that classifiers often confuse this class label with the other two class labels. Similar findings have also been reported in \cite{nguyen17damage}.  
For the disaster type task, the performance of the AIDR-DT model is higher compared to the DMD model. We observe that the DMD dataset is comparatively small, and the model is not performing well on the consolidated dataset. This characteristic is observed in other tasks as well. For the damage severity task, CrisisMMD is performing worse, which is also reflected in its dataset size, i.e., 2,493 images in the training set, as shown in Table \ref{tab:data_split_damage_severity}.
As expected, overall, for all tasks, the models with the consolidated datasets outperform individual datasets.

\begin{table}[h]
\centering
\caption{Results using different neural network models on the consolidated dataset with four different tasks. Trained and tested using the consolidated dataset. Comparable results are shown in \textbf{bold} and best results are shown in \underline{underlined}. IncepNet (InceptionNet), MobNet (MobileNet), EffiNet (EfficientNet)}
\label{tab:classification_results_net_comparison}
\scalebox{0.85}{
\begin{tabular}{@{}lrrrr|rrrr@{}}
\toprule
\multicolumn{1}{c}{\textbf{Arch}} & \multicolumn{1}{c}{\textbf{Acc}} & \multicolumn{1}{c}{\textbf{P}} & \multicolumn{1}{c}{\textbf{R}} & \multicolumn{1}{c}{\textbf{F1}} & \multicolumn{1}{c}{\textbf{Acc}} & \multicolumn{1}{c}{\textbf{P}} & \multicolumn{1}{c}{\textbf{R}} & \multicolumn{1}{c}{\textbf{F1}} \\  \midrule
\multicolumn{1}{c}{\textbf{\textbf{}}} & \multicolumn{4}{c}{\textbf{Disaster type}} & \multicolumn{4}{c}{\textbf{Informative}} \\ \cmidrule{2-5} \cmidrule{6-9}
ResNet18 & 0.790 & 0.783 & 0.790 & 0.785 & 0.852 & 0.851 & 0.852 & 0.851 \\
ResNet50 & 0.810 & 0.806 & 0.810 & \textbf{0.808} & 0.852 & 0.852 & 0.852 & 0.852 \\
ResNet101 & 0.817 & 0.812 & 0.817 & \textbf{0.813} & 0.853 & 0.853 & 0.853 & 0.852 \\
AlexNet & 0.756 & 0.756 & 0.756 & 0.754 & 0.827 & 0.829 & 0.827 & 0.828 \\
VGG16 & 0.800 & 0.796 & 0.800 & 0.798 & 0.859 & 0.858 & 0.859 & \textbf{0.858} \\
DenseNet(121) & 0.811 & 0.805 & 0.811 & \textbf{0.806} & 0.863 & 0.863 & 0.863 & \textbf{0.862} \\
SqueezeNet & 0.757 & 0.754 & 0.757 & 0.755 & 0.829 & 0.829 & 0.829 & 0.829 \\
InceptionNet (v3) & 0.562 & 0.609 & 0.562 & 0.528 & 0.663 & 0.723 & 0.663 & 0.593 \\
MobileNet (v2) & 0.785 & 0.781 & 0.785 & 0.782 & 0.850 & 0.849 & 0.850 & 0.849 \\
\rowcolor[HTML]{add8e6}
EfficientNet (b1) & 0.818 & 0.815 & 0.818 & \underline{\textbf{0.816}} & 0.864 & 0.863 & 0.864 & \underline{\textbf{0.863}} \\\cmidrule{2-5} \cmidrule{6-9}
\multicolumn{1}{c}{} & \multicolumn{4}{c}{\textbf{Humanitarian}} & \multicolumn{4}{c}{\textbf{Damage severity}} \\ \cmidrule{2-5} \cmidrule{6-9}
ResNet18 & 0.754 & 0.747 & 0.754 & 0.749 & 0.751 & 0.734 & 0.751 & 0.736 \\
ResNet50 & 0.770 & 0.762 & 0.770 & 0.762 & 0.763 & 0.746 & 0.763 & \textbf{0.751} \\
ResNet101 & 0.769 & 0.763 & 0.769 & \textbf{0.765} & 0.760 & 0.736 & 0.760 & 0.737 \\
AlexNet & 0.721 & 0.715 & 0.721 & 0.716 & 0.734 & 0.714 & 0.734 & 0.709 \\
VGG16 & 0.778 & 0.773 & 0.778 & \underline{\textbf{0.773}} & 0.769 & 0.750 & 0.769 & \textbf{0.753} \\
DenseNet(121) & 0.765 & 0.756 & 0.765 & 0.755 & 0.755 & 0.734 & 0.755 & 0.739 \\
SqueezeNet & 0.730 & 0.717 & 0.730 & 0.719 & 0.733 & 0.707 & 0.733 & 0.708 \\
InceptionNet (v3) & 0.598 & 0.637 & 0.598 & 0.509 & 0.660 & 0.623 & 0.660 & 0.615 \\
MobileNet (v2) & 0.751 & 0.745 & 0.751 & 0.746 & 0.746 & 0.727 & 0.746 & 0.730 \\
\rowcolor[HTML]{add8e6}
EfficientNet (b1) & 0.767 & 0.764 & 0.767 & \textbf{0.765} & 0.766 & 0.754 & 0.766 & \underline{\textbf{0.758}} \\ \bottomrule
\end{tabular}
}
\end{table}

\begin{table}[t]
\centering
\caption{Different neural network models with number of layer, parameters and memory requirement during the inference of  a binary (Informativeness) classification task.}
\label{tab:net_comparison_params}
\scalebox{0.85}{
\begin{tabular}{@{}lrrr@{}}
\toprule
\multicolumn{1}{@{}l}{\textbf{Model}} & \multicolumn{1}{r}{\textbf{\# Layer}} & \multicolumn{1}{r}{\textbf{\# Param (M)}} & \multicolumn{1}{r@{}}{\textbf{Memory (MB)}} \\ \midrule
ResNet18 & 18 & 11.18 & 74.61 \\
ResNet50 & 50 & 23.51 & 233.54 \\
ResNet101 & 101 & 42.50 & 377.58 \\
AlexNet & 8 & 57.01 & 222.24 \\
VGG16 & 16 & 134.28 & 673.87 \\
DenseNet (121) & 121 & 6.96 & 174.2 \\
SqueezeNet & 18 & 0.74 & 47.99 \\
InceptionNet (v3) & 42 & 24.35 & 206.01 \\
MobileNet (v2) & 20 & 2.23 & 8.49 \\
EfficientNet (b1) & 25 & 7.79 & 177.82 \\ \bottomrule
\end{tabular}
}
\end{table}

\begin{figure}[t]
\centering
\includegraphics[width=0.7\linewidth]{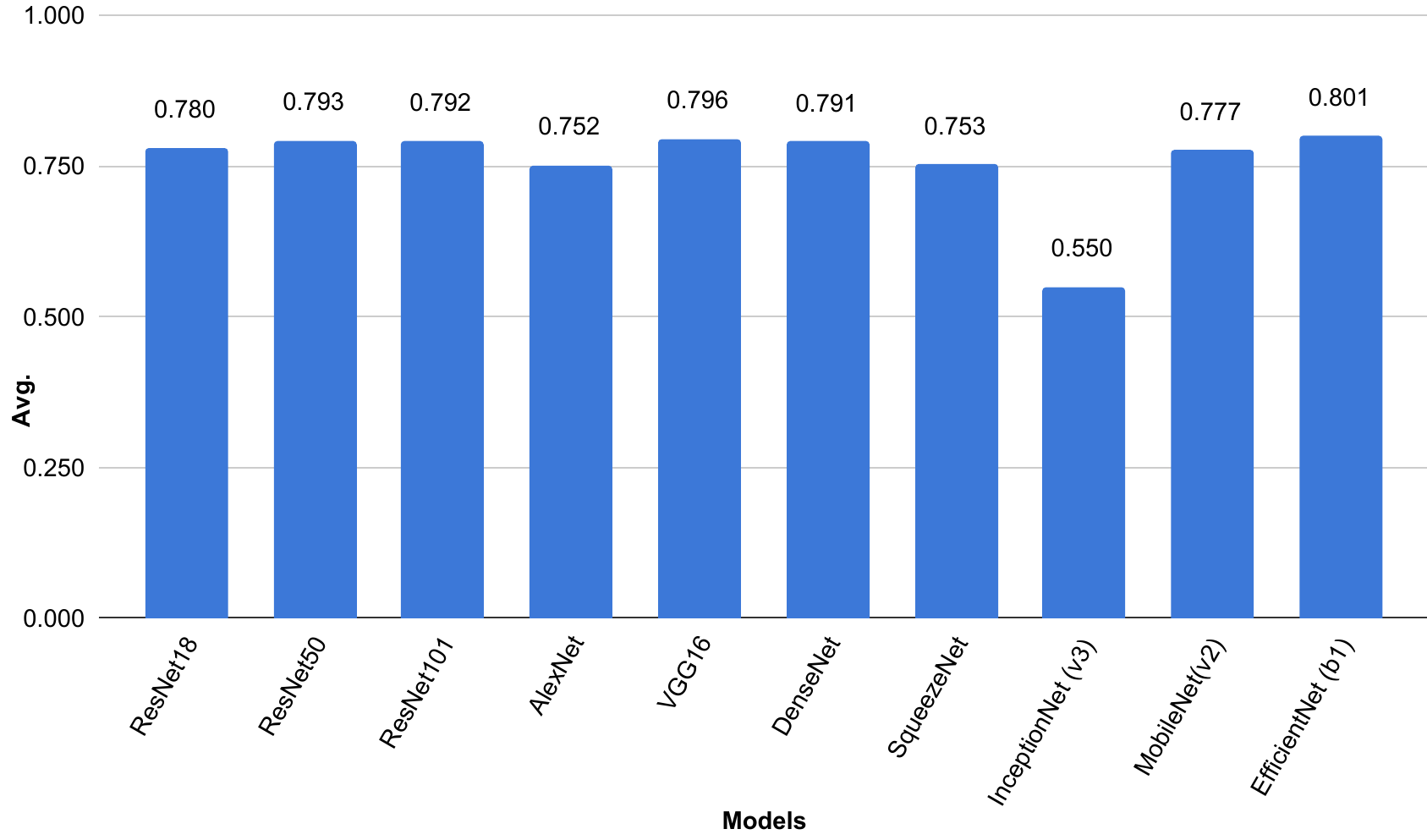}
\caption{Average F1 scores from all four tasks with different network architectures, which shows on average \textit{EfficientNet (b1)} performs better than other architectures.}
\label{fig:net_arch_f1_across_tasks}
\end{figure}

\begin{figure}[h]
        \centering
        \begin{subfigure}{.6\textwidth}
            \includegraphics[width=\textwidth]{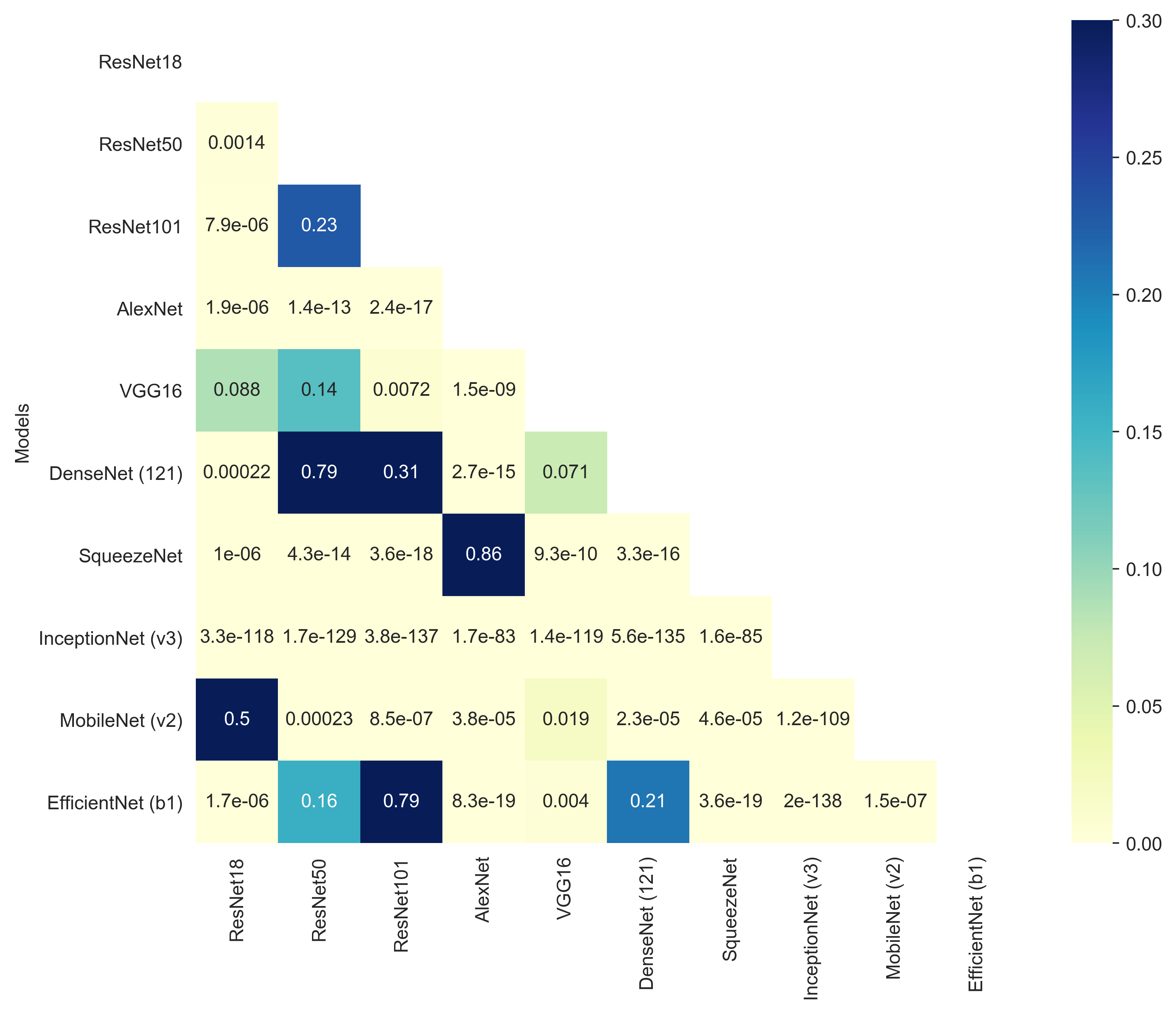}
            \caption{Disaster type}
        \end{subfigure}\\
        \begin{subfigure}{.6\textwidth}
                \includegraphics[width=\textwidth]{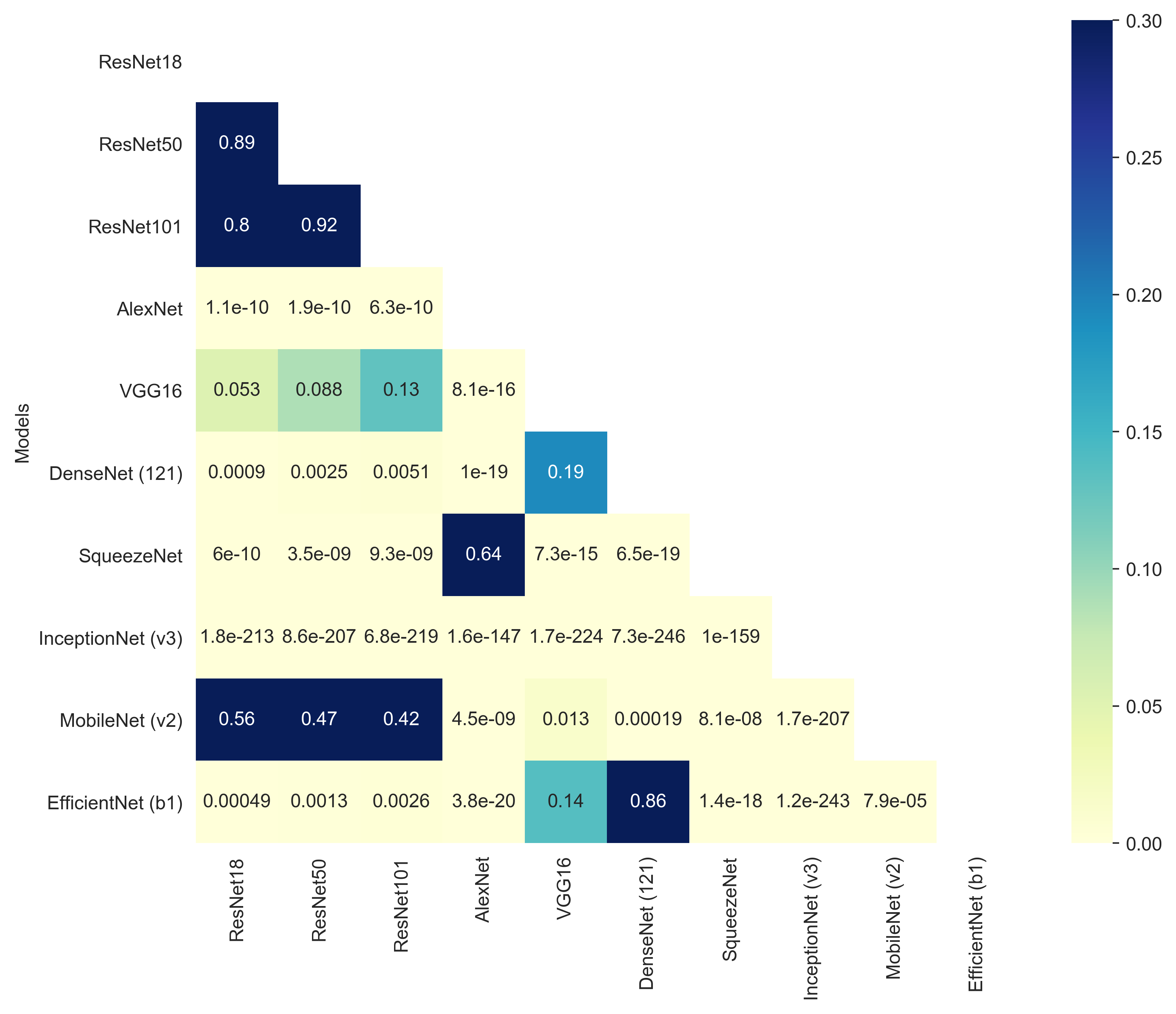}
                \caption{Informativeness}
        \end{subfigure}
    \caption{Statistical significant test among the different network architectures for \textit{Disaster type} and \textit{Informativeness} tasks. {\em$P$}-values are presented in cells. Light yellow color represent they are statistically significant with $p<0.05$}
    \label{fig:net_arch_statistical_significant_test_dt_info}
\end{figure}

\begin{figure}[h]
        \centering
        \begin{subfigure}{.65\textwidth}
                \includegraphics[width=\textwidth]{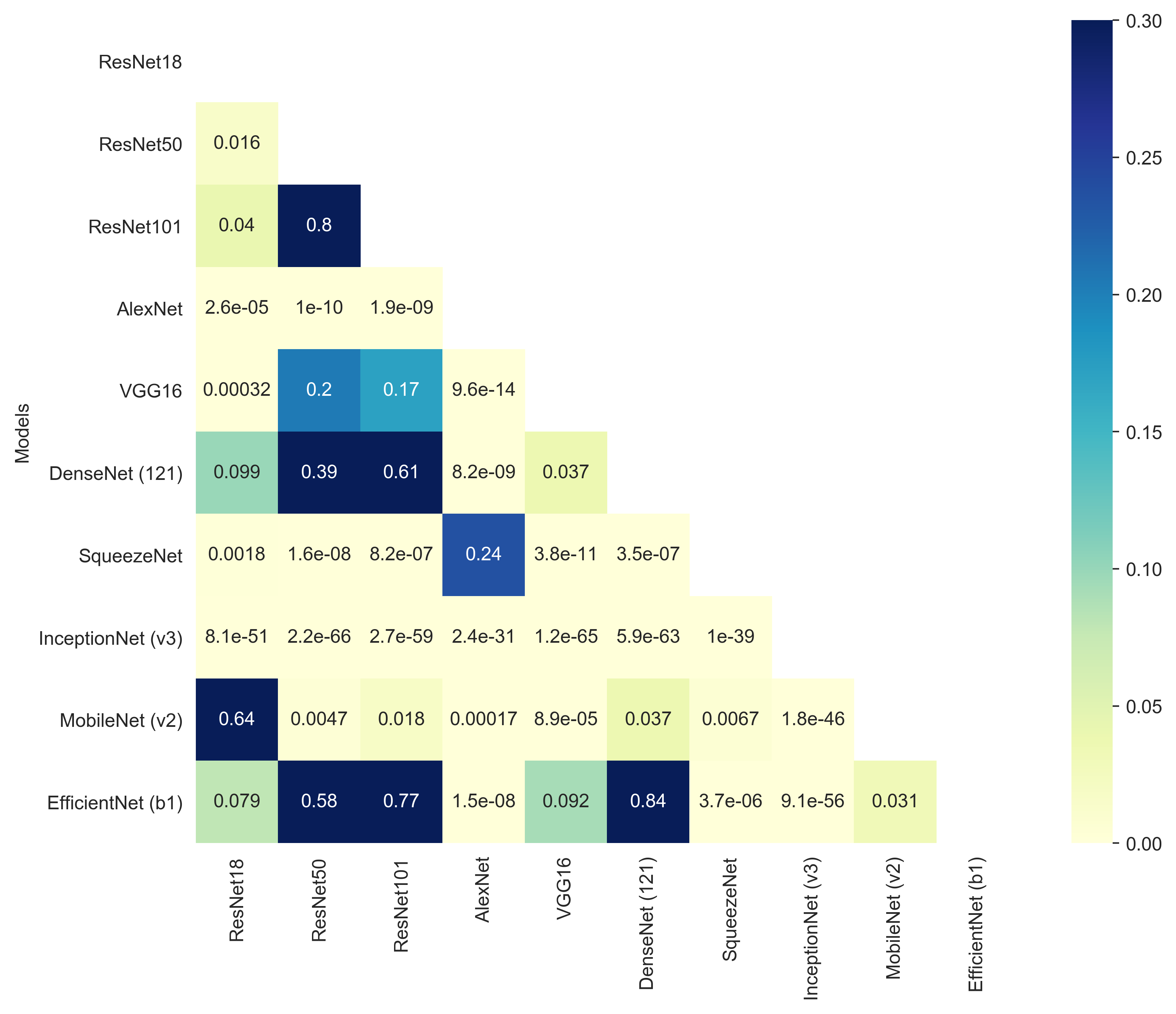}
                \caption{Humanitarian}
        \end{subfigure}
        \begin{subfigure}{.65\textwidth}
                \includegraphics[width=\textwidth]{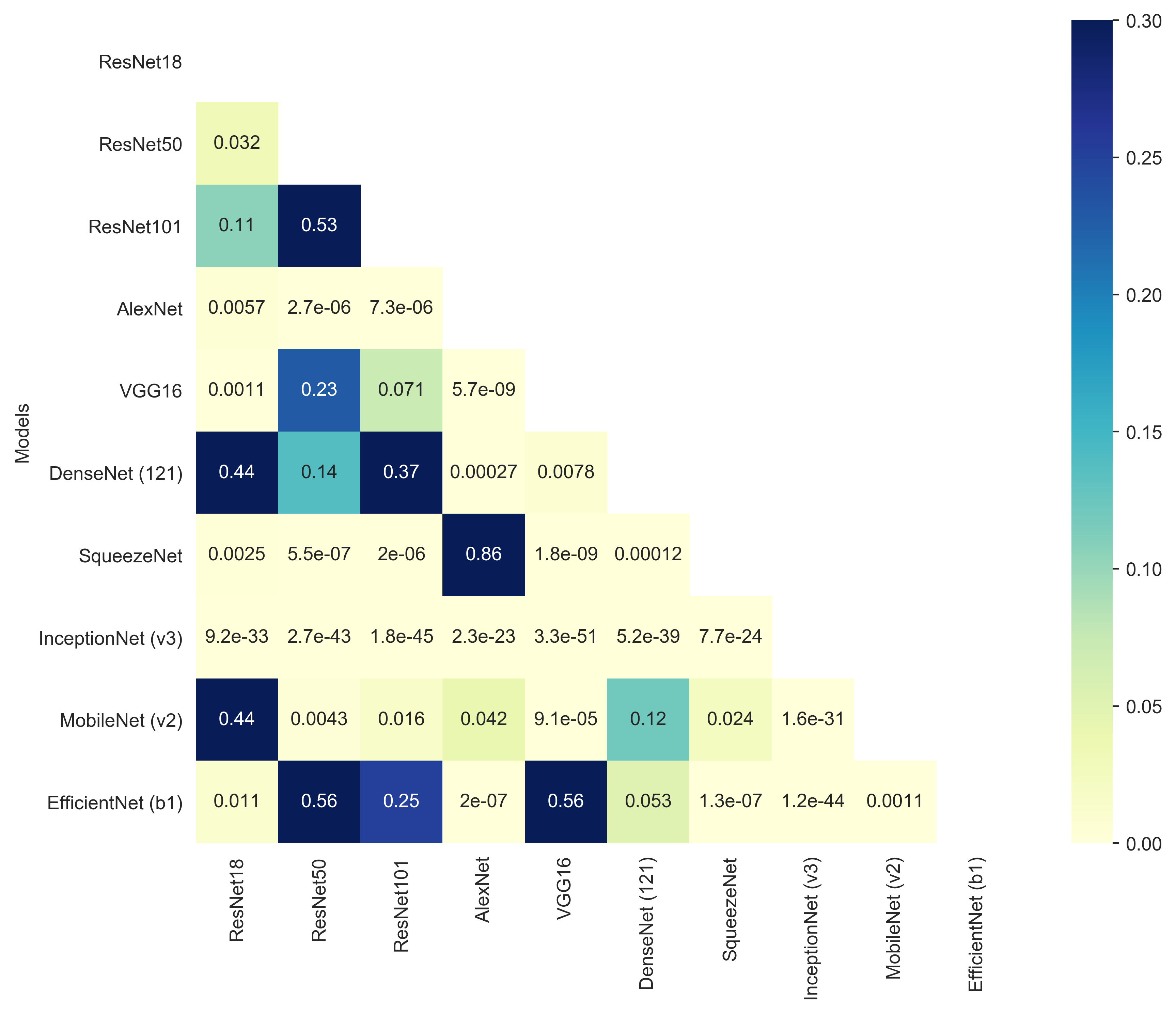}
                \caption{Damage severity}
        \end{subfigure}   
    \caption{Statistical significant test among the different network architectures for \textit{Humanitarian} and \textit{Damage severity} tasks. {\em$P$}-values are presented in cells. Light yellow color represent they are statistically significant with $p<0.05$}
    \label{fig:net_arch_statistical_significant_test_hum_ds}
\end{figure}

\subsection{Network Architectures Comparison}
\label{ssec:results_network_comparison}
In Table~\ref{tab:classification_results_net_comparison}, we report results using different network architectures on consolidated datasets for different tasks, i.e., trained and tested using a consolidated dataset. Across different tasks, EfficientNet (b1) is performing better than other models as shown in Figure \ref{fig:net_arch_f1_across_tasks}, except for humanitarian task, for which VGG16 is outperforming other models. Comparatively the second-best models are VGG16, ResNet50, ResNet101, and DenseNet (101). From the results of different tasks, we observe that InceptionNet (v3) is the worst-performing model. 

The performance difference among different models such as EfficientNet (b1), VGG16, ResNet50, ResNet101, and DenseNet (101) are low, hence, we have done statistical test to understand whether such small differences are significant. We used McNemar's test for binary classification task, (i.e., informativeness) and Bowker's test for other multiclass classification tasks. More details of this test can be found in \cite{HOFFMAN2019233}. We have done such tests between two models to see a pair-wise difference. In Figure \ref{fig:net_arch_statistical_significant_test_dt_info} and \ref{fig:net_arch_statistical_significant_test_hum_ds}, we report the results of significant tests. The value in the cell represent the \textit{\textbf{$P$}}-value and the light yellow color represent they are statistically significant with \textit{\textbf{$P<0.05$}}. From the Figure \ref{fig:net_arch_statistical_significant_test_dt_info}, we see that for disaster type task the \textit{\textbf{$P$}}-value is higher than $0.05$ in comparison between EfficientNet (b1) vs.\ ResNet50, ResNet101 and DenseNet (121), which clearly reflects among the results reported in Table \ref{tab:classification_results_net_comparison}. Similarly the difference is very low between EfficientNet (b1) vs.\  VGG16 and DenseNet (121). For humanitarian and damage severity tasks, we observed similar behaviors. By analyzing all four tasks it appears VGG16 is the second best performing model. 

In Table~\ref{tab:net_comparison_params}, we also report different neural network models with their number of layers, parameters, and memory consumption during the inference of informativeness task. There is usually a trade-off between the performance and computational complexity of different deep neural networks. In terms of memory consumption and the number of parameters, VGG16 seems expensive than others. Based on the performance and computational complexity, we can conclude that EfficientNet can be the best option for real-time applications. We computed throughput for EfficientNet using a batch size of 128, and it can process ${\sim}$260 images per second on an NVIDIA Tesla P100 GPU. ResNet18 is a reasonable choice among different ResNet models, given that its computational complexity is significantly less than other ResNet models. 

\begin{table}[h]
\centering
\caption{Results with data augmentation and weight decay using different neural network models on the consolidated dataset for all four tasks. \textit{\textbf{Diff.}} represents the difference without RandAugment results presented in Table \ref{tab:classification_results_net_comparison}. * represents statistically significant (with $P<0.05$) compared to the without RandAugment results.}
\label{tab:classification_results_data_augmentation}
\scalebox{0.85}{
\begin{tabular}{@{}lrrrrrrrrrr@{}}
\toprule
\multicolumn{1}{c}{\textbf{Arch}} &
  \multicolumn{1}{c}{\textbf{Acc}} &
  \multicolumn{1}{c}{\textbf{P}} &
  \multicolumn{1}{c}{\textbf{R}} &
  \multicolumn{1}{c}{\textbf{F1}} &
  \multicolumn{1}{c}{\textbf{Diff.}} &
  \multicolumn{1}{c}{\textbf{Acc}} &
  \multicolumn{1}{c}{\textbf{P}} &
  \multicolumn{1}{c}{\textbf{R}} &
  \multicolumn{1}{c}{\textbf{F1}} &
  \multicolumn{1}{c}{\textbf{Diff.}} \\ \midrule
\multicolumn{1}{c}{} & \multicolumn{5}{c}{\textbf{Disaster type}}   & \multicolumn{5}{c}{\textbf{Informative}}      \\\cmidrule{2-11}
ResNet18             & 0.812 & 0.807 & 0.812 & 0.809 & 2.4  & 0.848 & 0.847 & 0.848 & 0.847 & -0.4 \\
ResNet50             & 0.817 & 0.81  & 0.817 & 0.812 & 0.4  & 0.863 & 0.863 & 0.863 & 0.862 & 1.0  \\
\rowcolor[HTML]{add8e6}
ResNet101            & 0.819 & 0.815 & 0.819 & 0.816 & 0.3  & 0.857 & 0.858 & 0.857 & 0.858 & 0.6  \\
AlexNet              & 0.755 & 0.753 & 0.755 & 0.753 & -0.1 & 0.827 & 0.826 & 0.827 & 0.825 & -0.3 \\
VGG16                & 0.803 & 0.797 & 0.803 & 0.798 & 0.0  & 0.855 & 0.855 & 0.855 & 0.855 & -0.3 \\
DenseNet (121)       & 0.817 & 0.811 & 0.817 & 0.813 & 0.7  & 0.858 & 0.858 & 0.858 & 0.857 & -0.5 \\
SqueezeNet           & 0.726 & 0.719 & 0.726 & 0.717 & -3.8 & 0.821 & 0.820 & 0.821 & 0.820 & -0.9 \\
\rowcolor[HTML]{add8e6}
InceptionNet (v3)    & 0.808 & 0.801 & 0.808 & *0.802 & 25.4 & 0.860 & 0.859 & 0.860 & *0.859 & 33.1 \\
\rowcolor[HTML]{add8e6}
MobileNet (v2)       & 0.793 & 0.788 & 0.793 & 0.789 & 0.7  & 0.854 & 0.853 & 0.854 & 0.853 & 0.4  \\
\rowcolor[HTML]{add8e6}
EfficientNet (b1)    & 0.838 & 0.834 & 0.838 & \textbf{0.835} & 1.9  & 0.869 & 0.868 & 0.869 & \textbf{0.868} & 0.5  \\ \cmidrule{2-11}
\multicolumn{1}{c}{\textbf{}} &
  \multicolumn{5}{c}{\textbf{Humanitarian}} &
  \multicolumn{5}{c}{\textbf{Damage severity}} \\\cmidrule{2-11}
ResNet18             & 0.745 & 0.738 & 0.745 & 0.741 & -0.8 & 0.757 & 0.736 & 0.757 & 0.739 & 0.3  \\
ResNet50             & 0.774 & 0.769 & 0.774 & 0.768 & 0.6  & 0.763 & 0.745 & 0.763 & 0.749 & -0.2 \\
\rowcolor[HTML]{add8e6}
ResNet101            & 0.774 & 0.778 & 0.774 & 0.775 & 1    & 0.766 & 0.753 & 0.766 & 0.757 & 2.0  \\
AlexNet              & 0.718 & 0.709 & 0.718 & 0.709 & -0.7 & 0.728 & 0.712 & 0.728 & 0.713 & 0.4  \\
VGG16                & 0.772 & 0.766 & 0.772 & 0.767 & -0.6 & 0.767 & 0.748 & 0.767 & 0.752 & -0.1 \\
DenseNet (121)       & 0.759 & 0.756 & 0.759 & 0.755 & 0    & 0.760 & 0.741 & 0.760 & 0.747 & 0.8  \\
SqueezeNet           & 0.720 & 0.713 & 0.720 & 0.712 & -0.7 & 0.729 & 0.708 & 0.729 & 0.702 & -0.6 \\
\rowcolor[HTML]{add8e6}
InceptionNet (v3)    & 0.762 & 0.753 & 0.762 & *0.754 & 25.6 & 0.758 & 0.735 & 0.758 & *0.739 & 11.5 \\
\rowcolor[HTML]{add8e6}
MobileNet (v2)       & 0.759 & 0.749 & 0.759 & 0.751 & 0.5  & 0.758 & 0.737 & 0.758 & 0.738 & 0.8  \\
\rowcolor[HTML]{add8e6}
EfficientNet (b1)    & 0.785 & 0.784 & 0.785 & \textbf{0.784} & 1.9  & 0.777 & 0.762 & 0.777 & \textbf{*0.765} & 0.7  \\ \bottomrule
\end{tabular}
}
\end{table}

\begin{figure}[h]
        \centering
        \begin{subfigure}{.65\textwidth}
            \includegraphics[width=\textwidth]{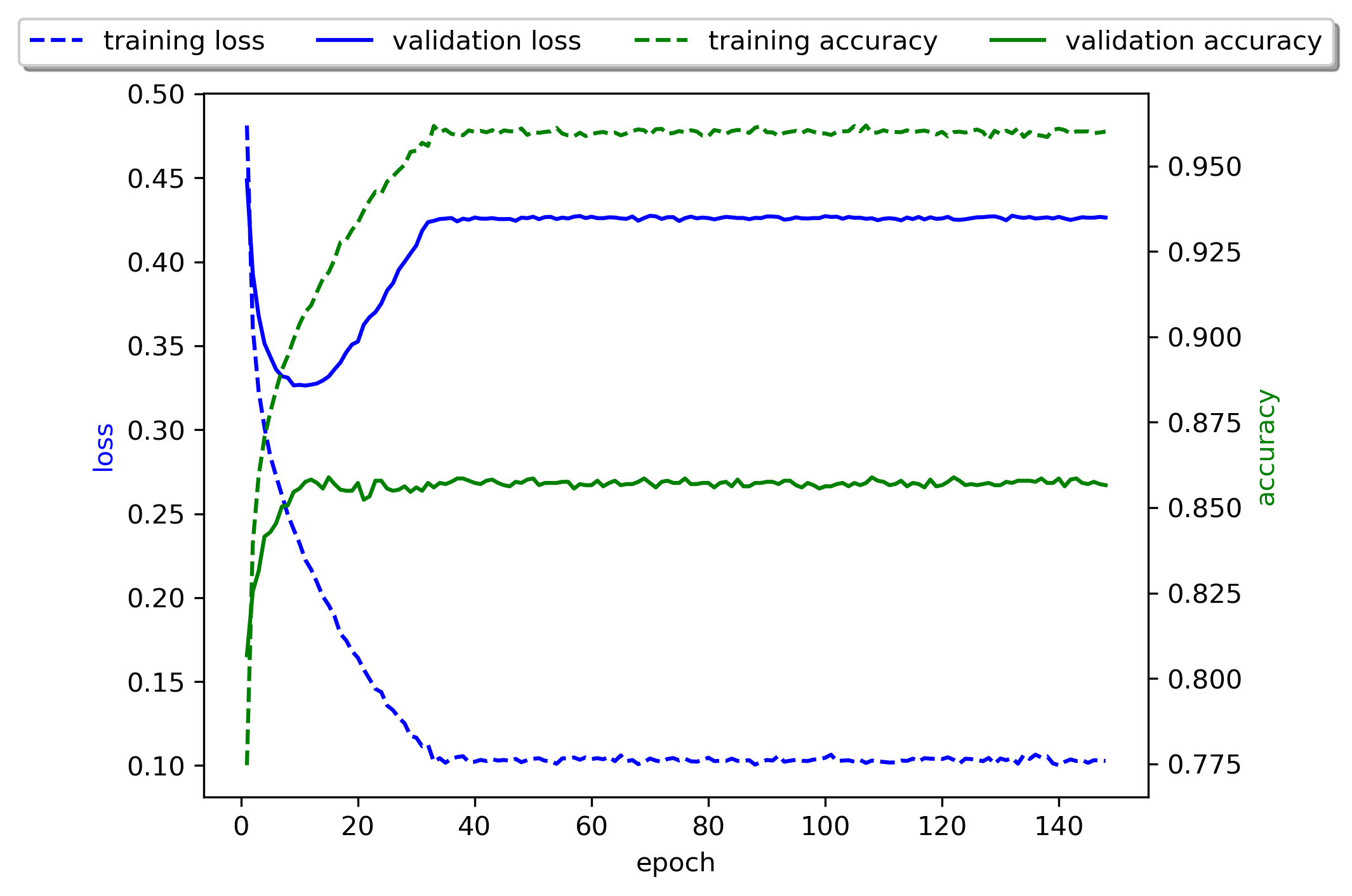}
            \caption{Without RandAugment}
            \label{fig:info_efficient_no_aug}
        \end{subfigure}\\
        \begin{subfigure}{.65\linewidth}
                \includegraphics[width=\textwidth]{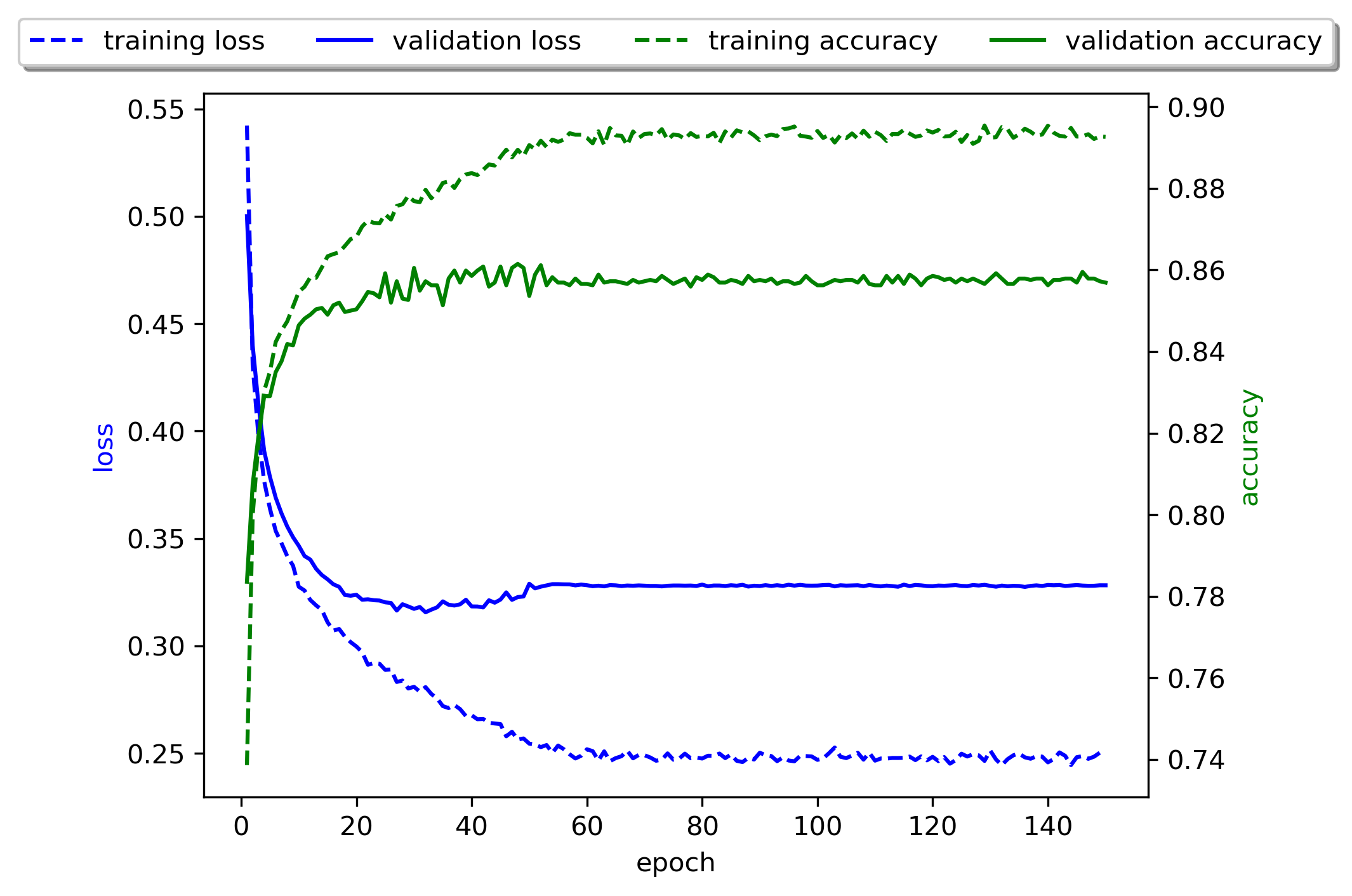}
                \caption{With RandAugment and weight decay}
                \label{fig:info_efficient_aug}
        \end{subfigure}
    \caption{Training/validation losses and accuracies without and with augmentation for \textit{\textbf{Informativeness}} task.}
    \label{fig:net_arch_statistical_significant_test_info}
\end{figure}

\begin{figure}[h]
        \centering
        \begin{subfigure}{.65\linewidth}
                \includegraphics[width=\textwidth]{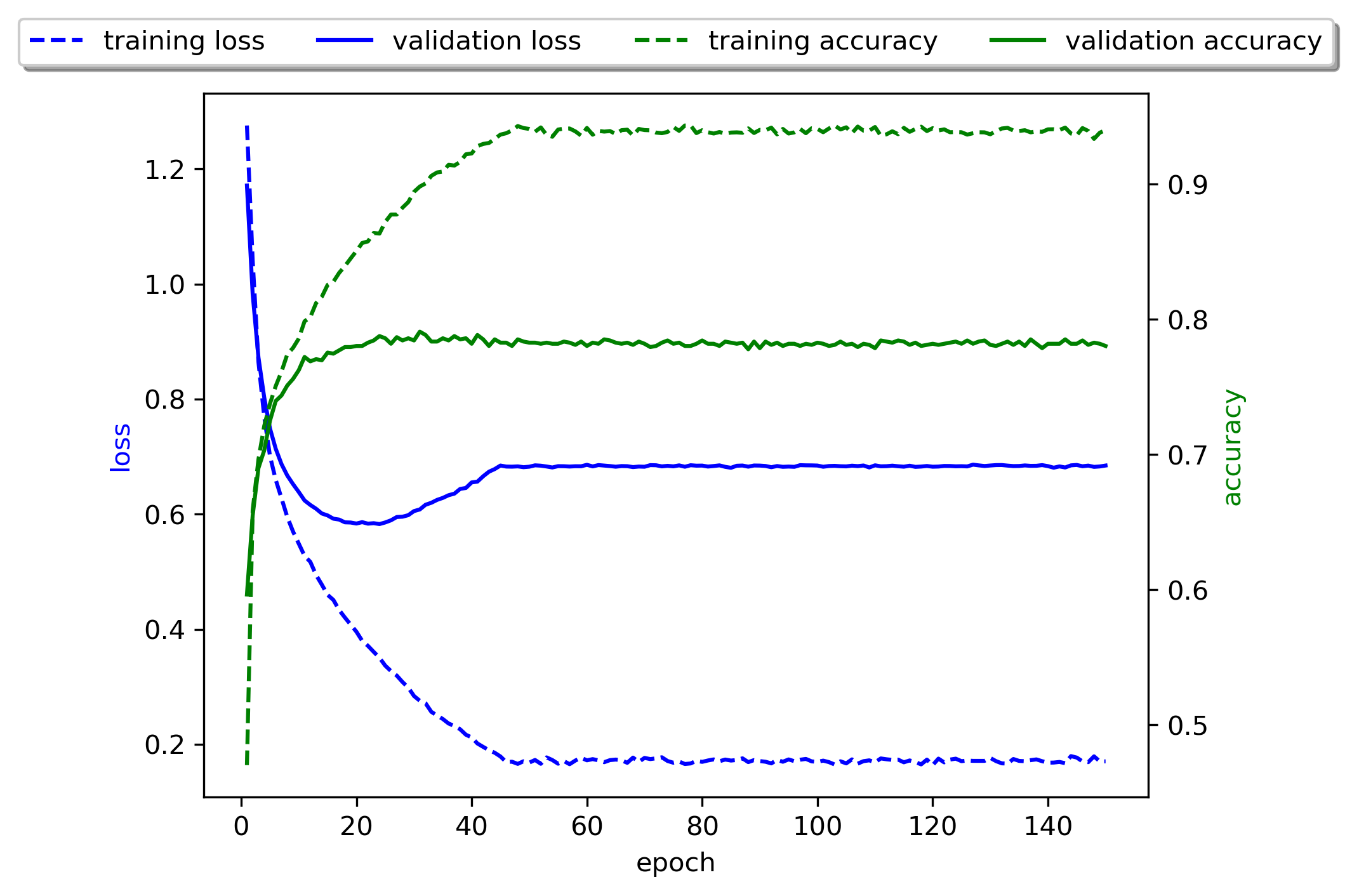}
                \caption{Without RandAugment}
                \label{fig:hum_efficient_no_aug}
        \end{subfigure}\\
        \begin{subfigure}{.65\linewidth}
                \includegraphics[width=\textwidth]{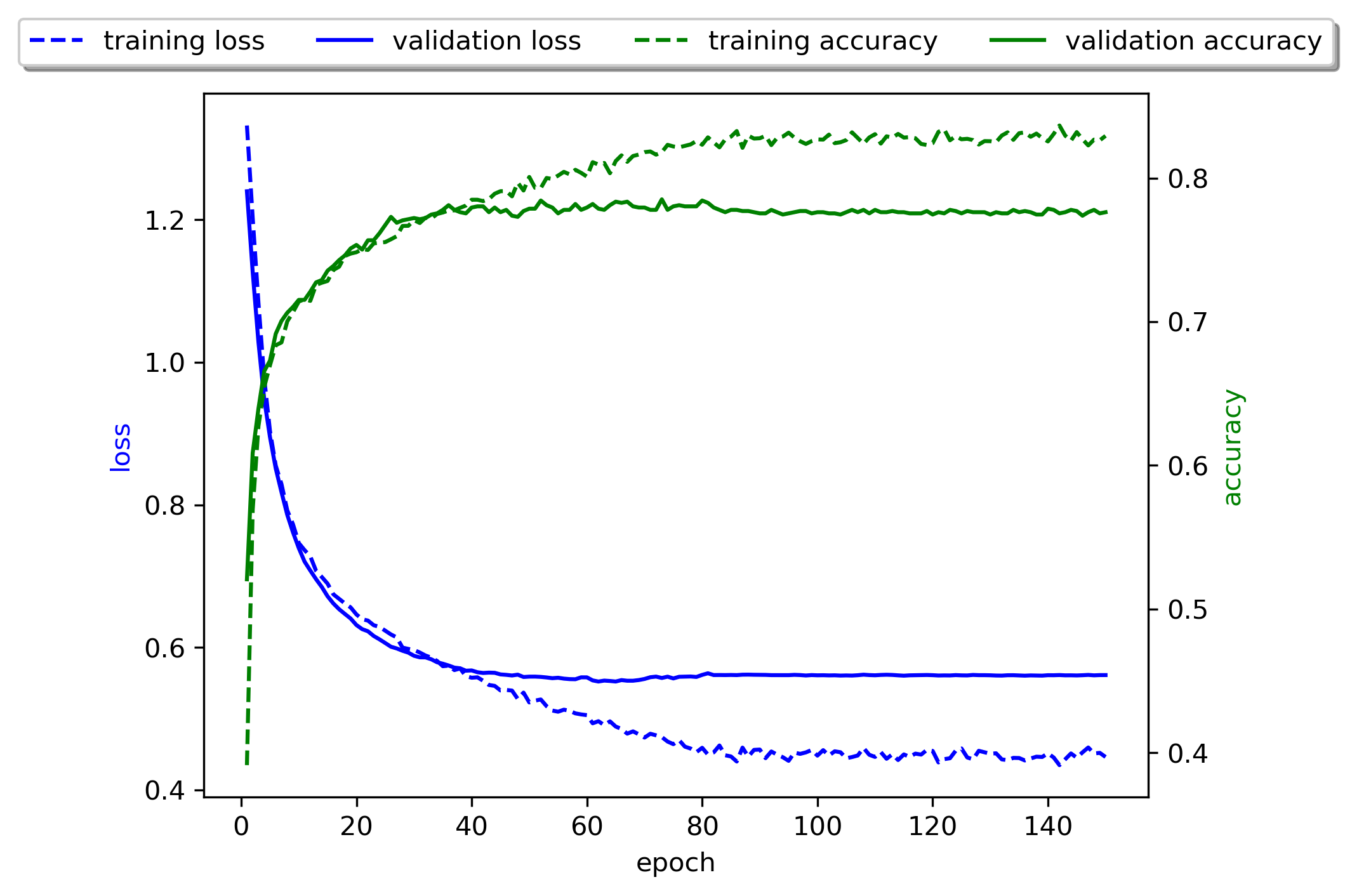}
                \caption{With RandAugment and weight decay}
                \label{fig:hum_efficient_aug}
        \end{subfigure}   
    \caption{Training/validation losses and accuracies without and with augmentation for \textit{\textbf{Humanitarian}} task.}
    \label{fig:net_arch_statistical_significant_test_hum}
\end{figure}

\subsection{Data Augmentation}
\label{ssec:results_data_augmentation}
To reduce the overfitting and to have more generalized models, we used data augmentation and weight decay. In Table \ref{tab:classification_results_data_augmentation}, we report the results for all tasks and using all network architectures. The column \textit{\textbf{Diff.}} report the difference between the results presented in Table \ref{tab:classification_results_net_comparison} where no RandAugment or \textit{weight decay} has been applied. The improved results are highlighted with light blue color for all tasks. Out of 40 experiments (10 network architectures \xmark~ 4 tasks), for 26 cases, the augmentation with weight decay improved the performances. 

On the improved cases, we also computed a statistical significance test between no RandAugment and RandAugment with \textit{weight decay} models. We found that the improvements for the models with InceptionNet (v3) are statistically significant in all tasks. Only the improved performance with EfficientNet (b1) for damage severity task is statistically significant, and for other tasks, they are not statistically significant. We investigated training and validation losses over the number of epochs. In Figure \ref{fig:net_arch_statistical_significant_test_info} and \ref{fig:net_arch_statistical_significant_test_hum}, we report training, validation losses and accuracies for EfficientNet (b1) model for Informativeness and Humanitarian tasks, respectively. From the figures \ref{fig:info_efficient_no_aug} and \ref{fig:hum_efficient_no_aug}, we clearly see that models are overfitting, whereas Figures \ref{fig:info_efficient_aug} and \ref{fig:hum_efficient_aug} show that models are more generalized. These findings demonstrate the benefits of augmentation and weight decay.

\subsection{Semi-supervised Learning}
\label{ssec:results_semisupervised}
In Table \ref{tab:classification_results_ns}, we present the results of the Noisy student-based self-training approach without/with RandAugment results. We have an $\sim1\%$ improvement for the \textit{Informativeness} task. For the \textit{Humanitarian} task, the performance is similar to RandAugment. For the \textit{Damage severity} task, the performance of Noisy student is the same as without RandAugment but lower than RandAugment. 

We postulate the following possible reasons for the lack of improvements in semi-supervised learning experiments:

\begin{enumerate}
    \itemsep0em
    
    \item Semi-supervised learning usually performs better when trained from scratch instead of fine-tuning from a pretrained model. This phenomenon is explored in \cite{zhou2018semi} where the authors reported the performance gained from semi-supervised learning methods are usually smaller when trained from a pretrained model. We could not train the student model from scratch as our labeled datasets are small, and it degrades performance even more.
    
    \item We had to use a much smaller labeled batch size of 16 compared to those used in \cite{xie2020self} (512 or higher) due to GPU constraints. Having a larger labeled batch size and, consequently, more unlabeled images in each batch may yield a better result. 
\end{enumerate}

\begin{table}[h]
\centering
\caption{Results with Noisy student self-training approach using \textit{Efficient (b1)} neural network models on the consolidated datasets for all four tasks. NS: Noisy student }
\label{tab:classification_results_ns}
\scalebox{0.85}{
\begin{tabular}{@{}lrrrr@{}}
\toprule
\multicolumn{1}{c}{\textbf{Exp.}} & \multicolumn{1}{c}{\textbf{Acc}} & \multicolumn{1}{c}{\textbf{P}} & \multicolumn{1}{c}{\textbf{R}} & \multicolumn{1}{c}{\textbf{F1}} \\ \midrule
 & \multicolumn{4}{c}{\textbf{Disaster type}} \\\midrule
Without RandAugment & 0.818 & 0.815 & 0.818 & 0.816 \\
RandAugment & 0.838 & 0.834 & 0.838 & 0.835 \\
NS &  0.793 & 0.812 & 0.793 & 0.794 \\\midrule
 & \multicolumn{4}{c}{\textbf{Informative}} \\\midrule
Without RandAugment & 0.864 & 0.863 & 0.864 & 0.863 \\
RandAugment & 0.869 & 0.868 & 0.869 & 0.868 \\
NS & 0.878 & 0.878 & 0.878 & \underline{\textbf{0.876}} \\\midrule
 & \multicolumn{4}{c}{\textbf{Humanitarian}} \\\midrule
Without RandAugment & 0.767 & 0.764 & 0.767 & 0.765 \\
RandAugment & 0.785 & 0.784 & 0.785 & 0.784 \\
NS & 0.783 & 0.786 & 0.783 & 0.783 \\\midrule
 & \multicolumn{4}{c}{\textbf{Damage severity}} \\\midrule
Without RandAugment & 0.766 & 0.754 & 0.766 & 0.758 \\
RandAugment & 0.777 & 0.762 & 0.777 & 0.765 \\
NS & 0.773 & 0.753 & 0.773 & 0.759 \\ \bottomrule
\end{tabular}
}
\end{table}

\begin{table}[h]
\centering
\caption{Results of multitask learning with \textbf{incomplete/missing} labels.}
\label{tab:classification_results_multitask_learning_all_tasks}
\scalebox{0.85}{
\begin{tabular}{@{}lrrrr@{}}
\toprule
\multicolumn{1}{c}{\textbf{Task}} & \multicolumn{1}{c}{\textbf{Acc}} & \multicolumn{1}{c}{\textbf{P}} & \multicolumn{1}{c}{\textbf{R}} & \multicolumn{1}{c}{\textbf{F1}} \\ \midrule
Disaster type & 0.647 & 0.657 & 0.647 & 0.637 \\
Informativeness & 0.727 & 0.735 & 0.727 & 0.726 \\
Humanitarian & 0.775 & 0.772 & 0.775 & 0.773 \\
Damage sevirity & 0.744 & 0.732 & 0.744 & 0.737 \\ \bottomrule
\end{tabular}
}
\end{table}

\begin{table}[]
\centering
\caption{Results of multitask learning with different tasks combinations and \textbf{complete labels}. DT: Disaster Type, Info: Informative, Hum: Humanitarian, DS: Damage Severity.}
\label{tab:classification_results_multitask_learning_task_comb}
\scalebox{0.85}{
\begin{tabular}{@{}lrrrr@{}}
\toprule
\multicolumn{1}{c}{\textbf{Task}} & \multicolumn{1}{c}{\textbf{Acc}} & \multicolumn{1}{c}{\textbf{P}} & \multicolumn{1}{c}{\textbf{R}} & \multicolumn{1}{c}{\textbf{F1}} \\ \midrule
\multicolumn{5}{c}{\textbf{Two tasks: Info and DS}} \\ \midrule
Informative & 0.855 & 0.856 & 0.855 & 0.855 \\
Damage Severity & 0.806 & 0.799 & 0.806 & 0.802 \\\midrule
\multicolumn{5}{c}{\textbf{Two tasks: Info and Hum}} \\\midrule
Informative & 0.817 & 0.816 & 0.817 & 0.816 \\
Humanitarian & 0.761 & 0.756 & 0.761 & 0.758 \\\midrule
\multicolumn{5}{c}{\textbf{Four tasks: DT, Info, Hum and DS}} \\\midrule
Disaster type & 0.781 & 0.768 & 0.781 & 0.772 \\
Informative & 0.920 & 0.921 & 0.920 & 0.920 \\
Humanitarian & 0.827 & 0.807 & 0.827 & 0.816 \\
Damage Severity & 0.772 & 0.750 & 0.772 & 0.759 \\ \bottomrule
\end{tabular}
}
\end{table}

\subsection{Multitask Learning}
\label{ssec:multitask_learning}
Since the \textit{Crisis Benchmark Dataset} has not been designed to address the multitask learning, we needed to re-split it as discussed in Section \ref{ssec:exp_multitask_learning}. This resulted two different settings: {\em(i)} incomplete/missing labels, and {\em(ii)} complete aligned labels. The incomplete/missing labels in multitask learning is a challenging problem, which we addressed using masking, i.e., for an unlabeled output, we are not computing loss for that particular task. In Table \ref{tab:classification_results_multitask_learning_all_tasks}, we report the results of multitask learning with missing labels where we address all tasks. 
We also investigated different task combinations where all labels are present. In Table \ref{tab:classification_results_multitask_learning_task_comb}, we report the results of different tasks combinations where they have complete aligned labels. For different task combinations, performances differ due to their data sizes, label distribution, and task settings. 
The results with multitask learning are not directly comparable with our single task setup. However, they can serve as a baseline for future studies.

\begin{figure}[h]
\centering
\includegraphics[width=0.7\textwidth]{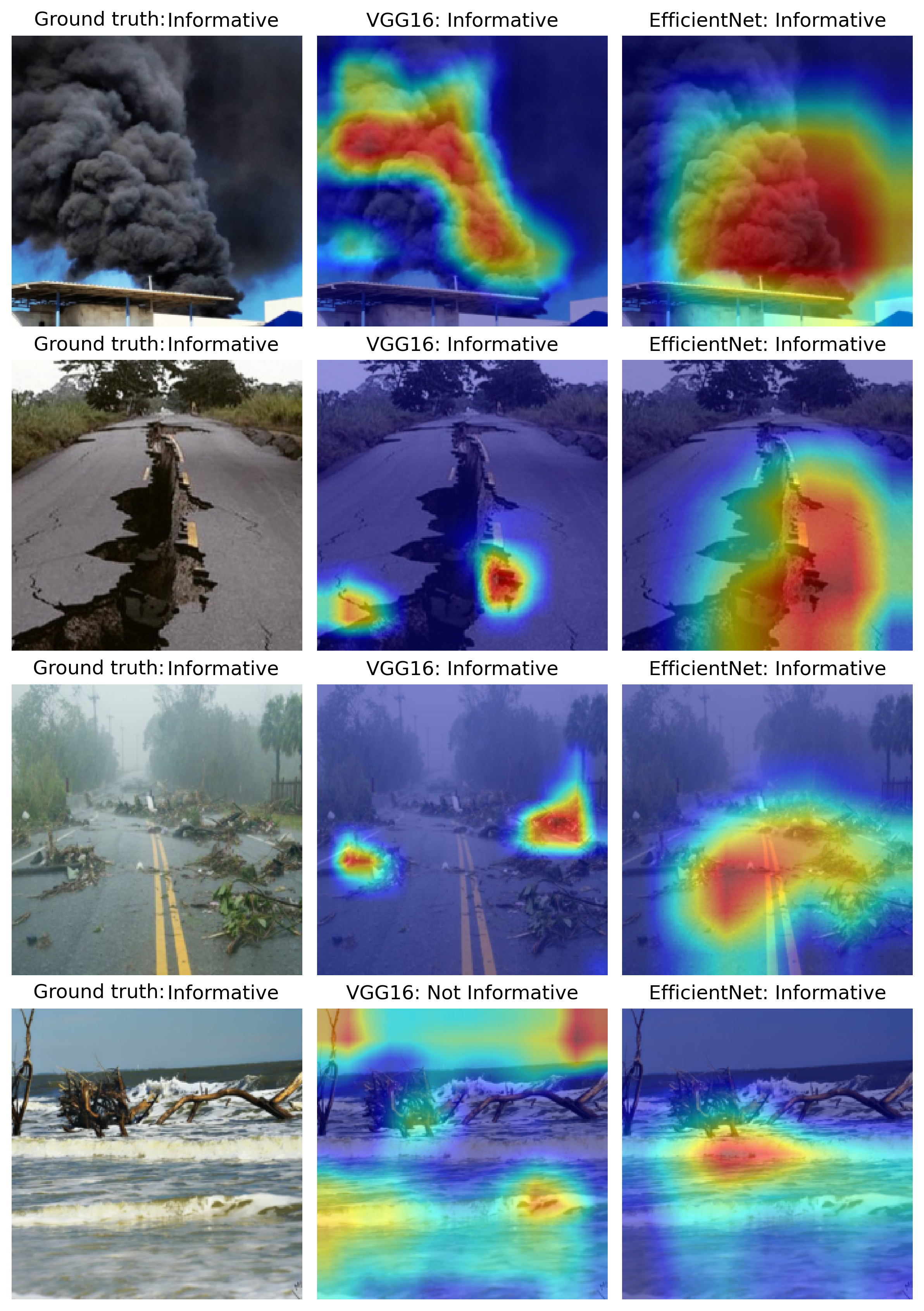}
\caption{GradCAM visualization of some images for the informativeness task.}
\label{fig:vis_info}
\end{figure}

\begin{figure}[h]
\centering
\includegraphics[width=0.7\textwidth]{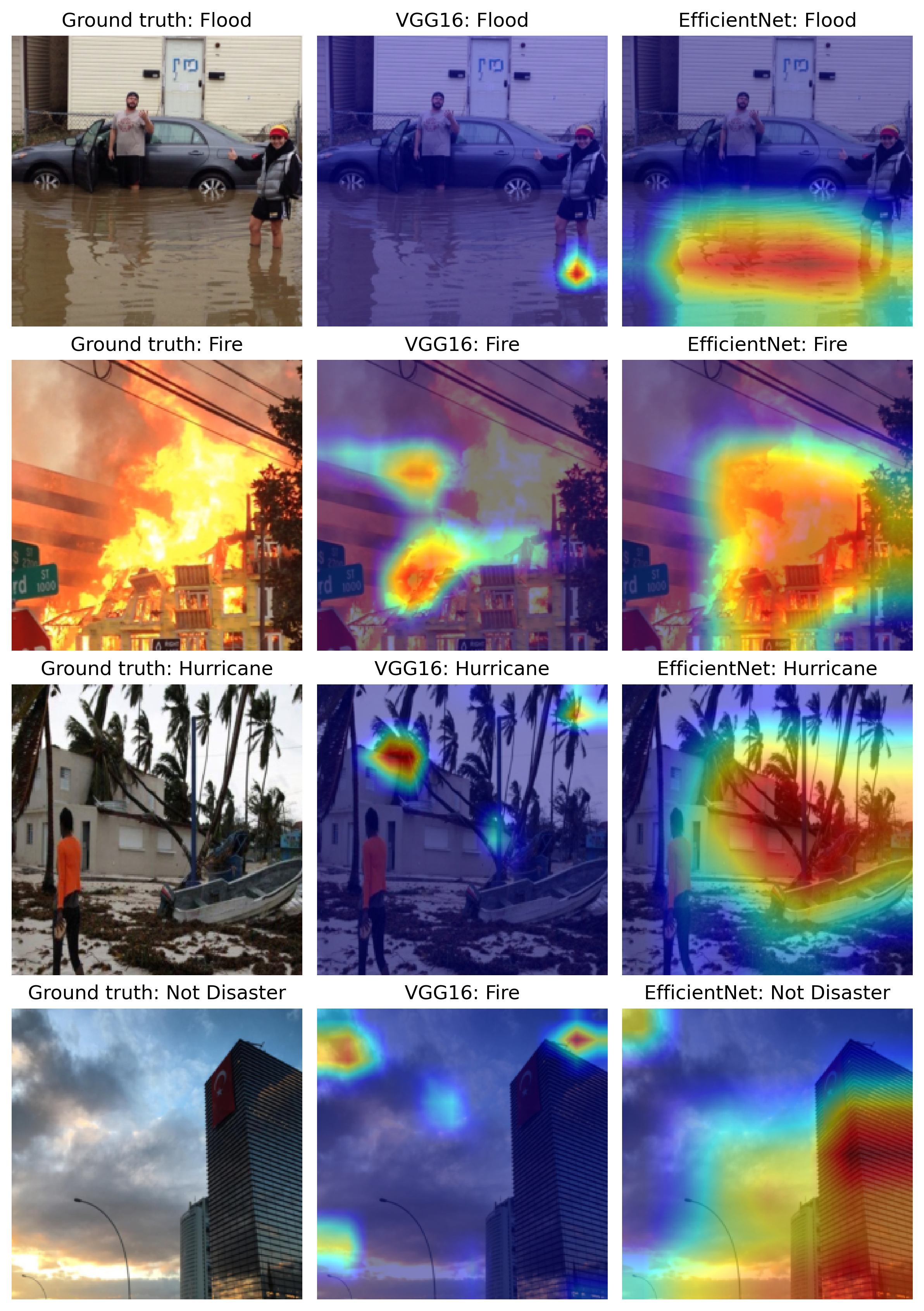}
\caption{Grad-CAM visualization of some images for the disaster type task.}
\label{fig:vis_dis}
\end{figure}

\subsection{Visual Explanation using Grad-CAM}
We explore how the neural networks arrive at their decision by utilizing Gradient-weighted Class Activation Mapping (Grad-CAM) \cite{selvaraju2017grad}. Grad-CAM uses the gradient of a target class flowing into the final convolution layer to produce a localization map highlighting the important regions in the image for that specific class. %
We report results for two candidate networks, i.e., VGG16 and EfficientNet, on two tasks, i.e., informativeness and disaster type. We use the models trained using RandAugment for this experiment. 

In Figure \ref{fig:vis_info}, we show the activation map for the predicted class for some images from the informativeness test set. From these images, it is apparent that EfficientNet performs better for localizing important regions in the image for the class of interest. VGG16 tends to depend on smaller regions for decision-making. The last row shows an image where VGG16 misclassified an informative image as not informative.

We show the activation map for some images from the test set of the disaster type task in Figure \ref{fig:vis_dis}. Here, the difference in localization quality between the two models is even more pronounced. The activation maps from VGG are difficult to interpret in the first and third images, even though the model classifies them correctly. The second image shows that VGG may focus on the smoke regions for classifying fire images. This explains why it identifies the last image as fire, misclassifying the clouds as smoke. 

Overall, these results suggest that EfficientNet does not only outperform other models in the numeric measures but it also produces activation maps that are easier to interpret. 
\section{Discussion and Future Work}
\label{sec:discussions}

\subsection{Our Findings}
Real-time event detection is an important problem from social media content. Our proposed pipeline and models are suitable to deploy them in real-time applications. The proposed models can also be used independently. For example, disaster type model can be used to monitor real-time disaster events. 

Our experiments were based on the research questions discussed in Section \ref{sec:introduction} below we report our findings based on them.

\textit{RQ1}: Our investigation to dataset comparison suggests that data consolidation helps, which answers our first research question. 

\textit{RQ2}: We also explore several deep learning models, which vary with performance and complexities. Among them, EfficientNet (b1) appears to be a reasonable option. Note that EfficientNet has a series of network architectures (b0-b7) and for this study, we only reported results with EfficientNet (b1). We aim to further explore other architectures. A small and low latency model is desired to deploy mobile and handheld embedded computer vision applications. The development of MobileNet~\cite{howard2017mobilenets} sheds light towards that direction. Our experimental results suggest that it is computationally simpler and provides a reasonable accuracy, only 2-3\% lower than the best models for different tasks. These findings answer out second research question.

\textit{RQ3}: We observe that strong data augmentation can improve performance, although this is not consistent across different tasks and models. Semi-supervised learning does not usually yield performance when trained using pretrained models and can sometimes even degrade it.

\textit{RQ4:} Multi-task learning can be an ideal solution for the real-time system as it can potentially provide speed-ups of multiple factors during inference. However, some tasks may perform worse than their single task settings in the presence of incomplete labels. Having aligned complete labels for different tasks can mitigate this issue.

\begin{table}[h]
\centering
\caption{\textbf{Recent relevant results reported in the literature}. \textit{\# C}: Number of class labels, Cls: Classification task, B: Binary, M: Multiclass, Incep: InceptionNet (v4), Info: Informativeness, Hum: Humanitarian, Event: Disaster event types, Infra.: Infrastructural damage, Severity: Severity Assessment. We converted some numbers from percentage (reported in the different literature) to decimal for an easier comparison.}
\label{tab:reported_results_in_literature}
\scalebox{0.7}{
\begin{tabular}{llrrllllrrrr}
\toprule
\multicolumn{1}{c}{\textbf{Ref.}} & \multicolumn{1}{c}{\textbf{Dataset}} & \multicolumn{1}{c}{\textbf{\# image}} & \multicolumn{1}{c}{\textbf{\# C}} & \multicolumn{1}{c}{\textbf{Cls.}} & \multicolumn{1}{c}{\textbf{Task}} & \multicolumn{1}{c}{\textbf{Models}} & \multicolumn{1}{c}{\textbf{Data Split}} & \multicolumn{1}{c}{\textbf{Acc}} & \multicolumn{1}{c}{\textbf{P}} & \multicolumn{1}{c}{\textbf{R}} & \multicolumn{1}{c}{\textbf{F1}} \\ \midrule
\cite{multimodalbaseline2020} & CrisisMMD & 12,708 & 2 & B & Info & VGG16 & Train/dev/test & 0.833 & 0.831 & 0.833 & 0.832 \\
\cite{multimodalbaseline2020} & CrisisMMD & 8,079 & 5 & M & Hum & VGG16 & Train/dev/test & 0.768 & 0.764 & 0.768 & 0.763 \\
\cite{Mouzannar2018} & DMD & 5879 & 6 & M & Event & Incep & 4 folds CV & 0.840 & - & - & - \\
\cite{agarwalcrisis} & CrisisMMD & 18,126 & 2 & B & Info & Incep & 5 folds CV & - & 0.820 & 0.820 & 0.820 \\
\cite{agarwalcrisis} & CrisisMMD & 18,126 & 2 & B & Infra. & Incep & 5 folds CV & - & 0.920 & 0.920 & 0.920 \\
\cite{agarwalcrisis} & CrisisMMD & 18,126 & 3 & B & Severity & Incep & 5 folds CV & - & 0.950 & 0.940 & 0.940 \\
\cite{abavisani2020multimodal} & CrisisMMD & 11,250 & 2 & B & Info & DenseNet & Train/dev/test & 0.816 & - & - & 0.812 \\
\cite{abavisani2020multimodal} & CrisisMMD & 3,359 & 5 & B & Hum & DenseNet & Train/dev/test & 0.834 & - & - & 0.870 \\
\cite{abavisani2020multimodal} & CrisisMMD & 3,288 & 3 & B & Severity & DenseNet & Train/dev/test & 0.629 & - & - & 0.661 \\ \bottomrule
\end{tabular}
}
\end{table}

\subsection{Comparison with the State of the Art}
We compared our results with recent and related state-of-the-art results, reported in Table \ref{tab:reported_results_in_literature}. However, it is not possible to have an end-to-end comparison for a few possible reasons: {\em(i)} different datasets and sizes -- see the second and third columns in Table \ref{tab:reported_results_in_literature}, {\em(ii)} different data splits (train/dev/test \textit{vs.} Cross Validation (CV) fold) even using same dataset -- see the \textit{Data Split} column in the same Table, {\em(iii)} different evaluation measures such as weighted P/R/F1-measure (first two rows) \cite{multimodalbaseline2020} \textit{vs.} accuracy (third row) \cite{Mouzannar2018} \textit{vs.} CV fold (fourth to sixth rows -- unspecified in \cite{agarwalcrisis} whether measures are macro, micro or weighted).

Even if they are not exactly comparable, we observe that on informativeness and humanitarian tasks, previously reported results (weighted F1) are 0.832 and 0.763, respectively, using the CrisisMMD dataset \cite{multimodalbaseline2020}. The authors in \cite{Mouzannar2018} reported a test accuracy of $0.840\pm0.0172$ for six disaster types tasks using the DMD dataset with a five-fold cross-validation run. The study in \cite{agarwalcrisis} report an F1 of 0.820 for informativeness, 0.920 for infrastructure damage, and 0.940 for damage severity. 
In another study, using the CrisisMMD dataset, authors report weighted-F1 of 0.812 and 0.870 for informativeness and humanitarian tasks, respectively~\cite{abavisani2020multimodal}. They used a small subset of the whole CrisisMMD dataset in their study. 
From the Table \ref{tab:reported_results_in_literature} we observe that the F1 for informativeness task ranges from 0.812 to 0.832 across studies, for humanitarian task it varies from 0.763 to 0.870, and for damage severity it varies from 0.661 to 0.940. Compared to them our best results (weighted F1) for disaster types, informativeness, humanitarian and damage severity are 0.835, 0.876, 0.784, and 0.765, respectively, on the consolidated single task dataset.

\subsection{Future Work}
As for future work we foresee several interesting research avenues. {\em(i)} Further exploration of semi-supervised learning to leverage a large amount of unlabeled social media data and address the limitations highlighted in Section \ref{ssec:results_semisupervised}. We believe addressing such limitations can help to advance state of the art. {\em(ii)} In multitask setup, one possible research direction is to address the problem of incomplete/missing labels, and the other is manually labeling \textit{Crisis Benchmark Dataset} for incomplete labels for all tasks. Both approaches will give the community grounds to explore multitask learning for real-time social media image classification. 

\section{Applications}
\label{sec:applications}
There are many application scenarios of the proposed models, however, in this section we discuss the ones that are highly relevant for crisis responders in humanitarian organizations. 

\noindent\textbf{Information for Situational Awareness:}
The information posted on social media during natural or human-induced disasters varies greatly. Studies have revealed that a big proportion of social media data consists of irrelevant information that is not useful for any kind of relief operations. For the decision-making process, humanitarian organizations are interested to have concise information about the ongoing situation to be aware of the event. The proposed models can help in filtering and reducing irrelevant content and provide a concrete summary.

\noindent\textbf{Actionable Information:}
Depending on their roles and mandate, humanitarian organizations differ in terms of their information needs. Several rapid response and relief agencies look for fine-grained information about specific incidents, which is also actionable. Such information types include reports of injured or dead people, critical infrastructure damage (e.g., a collapsed bridge), and rescue demand among others. Our study focused on coarse (i.e., binary) to fine-grained labels while also addressed four different but related tasks. Applications can be developed on top of our models, which can provide critical humanitarian information needs in crisis situations.

\noindent\textbf{Real-time Crisis Event Detection:} The proposed models (i.e., disaster type) can be deployed in real-time to continuously monitor social media and detect emergent events (e.g., fire, flood) around the world.

\section{Conclusions}
\label{sec:conclutions}
The imagery and textual content available on social media have been used by humanitarian organizations in times of disaster events. There has been limited work for disaster response image classification tasks compared to text. In this study, we posed four research questions and performed extensive experiments on four tasks such as disaster type, informativeness, humanitarian, and damage severity to answer those questions. Our experimental results on individual and consolidated datasets suggest that data consolidation helps. We investigated four tasks using various state-of-the-art neural network architectures and reported the best-performing models. The findings on data augmentation suggest that a more generalized model can be obtained with such approaches. Our investigation on semi-supervised and multitask learning suggests new research directions for the community. We also provide some insights of activation maps to demonstrate what class-specific information is learned by the network.

\section*{Funding} Not applicable.
\section*{Compliance with ethical standards}
\paragraph{Conflict of interest}
We have no conflicts of interest or competing interests to declare.
\paragraph{Availability of data and material}
The data used in this study are available at \url{https://crisisnlp.qcri.org/crisis-image-datasets-asonam20}.

\bibliographystyle{spmpsci}
\bibliography{main}

\end{document}